\documentclass{article}

\usepackage{microtype}
\usepackage{graphicx}
\usepackage{subcaption}
\usepackage{booktabs} 

\usepackage{xcolor}
\usepackage[normalem]{ulem}

\usepackage{float}         

\usepackage{multirow} 
\usepackage{booktabs}  
\usepackage{siunitx}   

\usepackage{tabularx}
\usepackage{array}
\usepackage{enumitem}

\usepackage[most]{tcolorbox}

\usepackage{hyperref}

\usepackage[preprint]{icml2026}

\makeatletter
\renewcommand{\algorithmiccomment}[1]{\unskip\hspace{0.6em}{\footnotesize\itshape$\triangleright$~#1}}
\makeatother

\usepackage{amsmath}
\usepackage{amssymb}
\usepackage{mathtools}
\usepackage{amsthm}

\usepackage[capitalize,noabbrev]{cleveref}

\theoremstyle{plain}

\theoremstyle{definition}

\theoremstyle{remark}

\usepackage[textsize=tiny]{todonotes}

\icmltitlerunning{\textsc{MAC}: Multi-Agent Constitution Learning}

\begin{document}

\twocolumn[
\icmltitle{\textsc{MAC}: Multi-Agent Constitution Learning}




  \icmlsetsymbol{equal}{*}


\begin{icmlauthorlist}
  \icmlauthor{Rushil Thareja}{mbzuai}
  \icmlauthor{Gautam Gupta}{iiitd}
  \icmlauthor{Francesco Pinto}{gdm}
  \icmlauthor{Nils Lukas}{mbzuai}
\end{icmlauthorlist}

\icmlaffiliation{mbzuai}{Mohamed bin Zayed University of Artificial Intelligence (MBZUAI)}
\icmlaffiliation{iiitd}{Indraprastha Institute of Information Technology Delhi (IIIT Delhi)}
\icmlaffiliation{gdm}{Google DeepMind}

\icmlcorrespondingauthor{Rushil Thareja}{rushil.thareja@mbzuai.ac.ae}



  \icmlkeywords{Machine Learning, ICML}

  \vskip 0.3in
]



\printAffiliationsAndNotice{}  

\begin{abstract}

Constitutional AI is a method to oversee and control LLMs based on a set of rules written in natural language. 
These rules are typically written by human experts, but could in principle be learned automatically given sufficient training data for the desired behavior.
Existing LLM-based prompt optimizers attempt this but are ineffective at learning constitutions since (i) they require many labeled examples and (ii) lack structure in the optimized prompts, leading to diminishing improvements as prompt size grows.
To address these limitations, we propose Multi-Agent Constitutional Learning (MAC), which optimizes over structured prompts represented as sets of rules using a network of agents with specialized tasks to accept, edit, or reject rule updates.
We also present MAC+, which improves performance by training agents on successful trajectories to reinforce updates leading to higher reward.
We evaluate MAC on tagging Personally Identifiable Information (PII), a classification task with limited labels where interpretability is critical, and demonstrate that it generalizes to other agentic tasks such as tool calling. 
MAC outperforms recent prompt optimization methods by over 50\%, produces human-readable and auditable rule sets, and achieves performance comparable to supervised fine-tuning and GRPO without requiring parameter updates.

\begin{center}
\href{https://github.com/rushil-thareja/MAC-Multi-Agent-Constitution-Learning}{Github Repo}
\;\;|\;\;
\href{https://pypi.org/project/mac-prompt/}{PyPI Package}
\;\;|\;\;
\href{https://www.mac-prompt.com/}{Website}
\end{center}


\end{abstract}
\section{Introduction}
Constitutional AI guides LLM behavior using structured prompts that enumerate natural-language rules defining what the model should and should not do \cite{bai2022constitutional, huang2024collective, sharma2025constitutional}. 
Because these rules are applied at inference time rather than embedded into model weights, they can be inspected, modified, and audited without retraining. 
This makes constitutional AI particularly attractive for domains such as healthcare, finance, and legal compliance, where transparency is essential and task definitions evolve over time.

\begin{figure}[t]
    \centering
    \includegraphics[width=0.98\columnwidth]{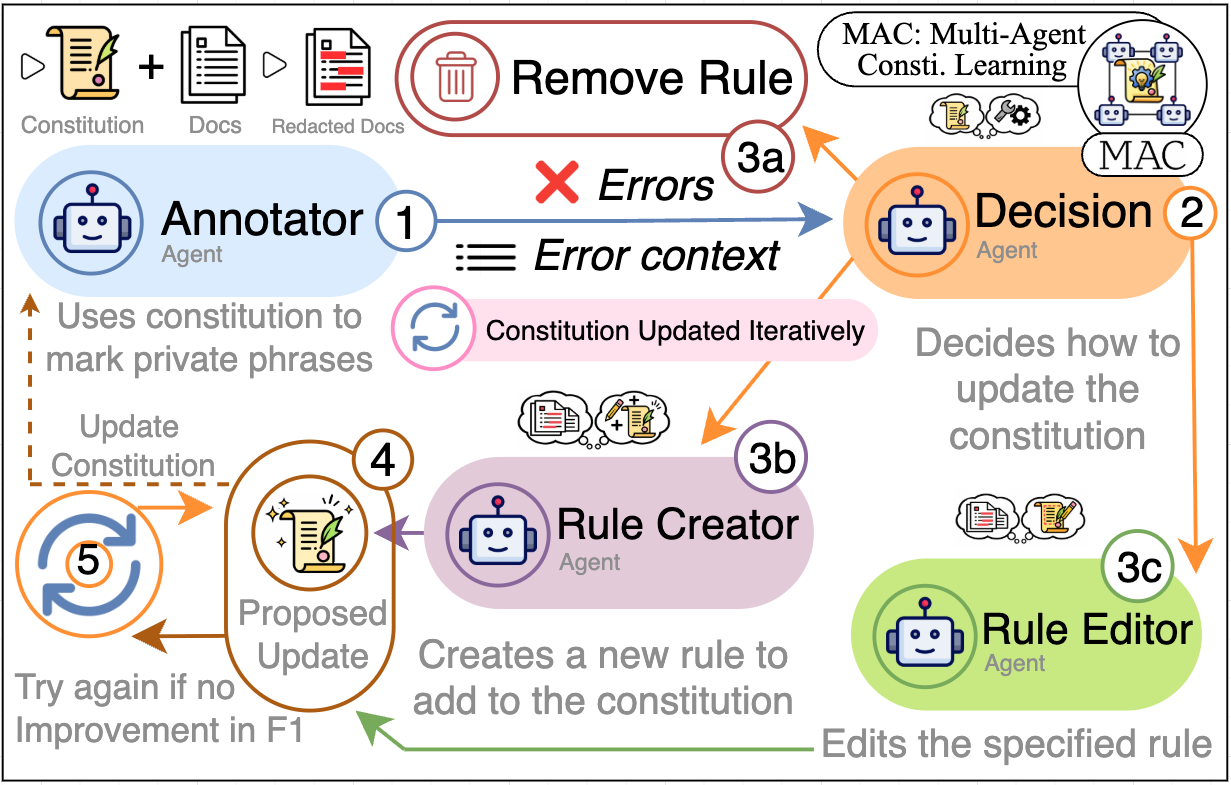}    
    \caption{In our MAC framework (1) an Annotator agent performs the downstream task given the constitution. (2) A Decision agent analyzes errors and reviews the constitution to add, edit, or remove rules. (3a) The Rule Proposer creates new candidate rules and (3b) the Rule Editor updates existing rules, producing a candidate constitution. (4) The update is evaluated with the task metric, and (5) the process repeats k times until performance improves.}
    \label{fig:eye-catcher}
    \label{support_counts_fig}
\end{figure}

A rule specifies when an action should or should not be taken. 
An effective rule must be both unambiguous to a human auditor and reliably followed by the LLM. 
However, writing such rules is difficult. 
Natural language that seems clear to humans can be interpreted inconsistently by models, and effectiveness must be validated empirically due to the probabilistic nature of LLMs.
In practice, developing a robust constitution requires iterative refinement where experts write rules, test them, observe failures, and revise. 
This process is slow, expensive, and must be repeated whenever task definitions change or new domains arise.

Automatic Prompt Optimization (APO)~\cite{yang2023large, GEPA} methods could in principle automate this process, but existing approaches are poorly suited for constitution learning. 
They typically require large amounts of labeled data, produce unstructured prompt modifications that are difficult to interpret, and show diminishing returns as prompts grow longer \cite{ramnath2025systematic}.
Moreover, they discard the structured, rule-based format that makes constitutional AI auditable in the first place.

We formalize \textit{constitution learning} as the task of automatically deriving a set of interpretable rules from limited input-output examples. 
Unlike general prompt optimization, constitution learning restricts updates to an explicit set of natural-language rules with predefined structure, preserving the transparency that makes constitutional AI valuable.

We propose Multi-Agent Constitutional Learning (MAC), which optimizes constitutions through a network of specialized agents that propose, critique, and refine rule updates.
Rather than optimizing a monolithic prompt, MAC treats the constitution as a modular structure where rules can be added, edited, or removed independently. 
Figure~\ref{tab:constitution_rules} illustrates our overall approach. 
The multi-agent design decomposes the optimization problem into subtasks, improving reliability and enabling lower-capacity models to contribute effectively \cite{chan2023chateval,chen2024reconcile,du2023improving}.

We evaluate MAC on Personally Identifiable Information (PII) tagging across finance \cite{li2025papillon}, legal \cite{echr_dataset}, and healthcare datasets \cite{maccrobat2020}. 
PII tagging is a natural testbed for constitution learning because it requires transparent decision-making, suffers from limited labeled data, and definitions vary substantially across domains and jurisdictions. 
Fine-tuning is also impractical in this setting due to cost, and inference-time rule application avoids the memorization risks associated with training on sensitive data \cite{carlini2023quantifyingmemorizationneurallanguage, huang2022largepretrainedlanguagemodels, kim2023propileprobingprivacyleakage}.
We further demonstrate that MAC generalizes beyond classification to agentic tasks by adapting it to tool calling on BFCL \cite{patil2025bfcl}.

\subsection{Contributions}
We make the following contributions:
\begin{itemize}
\item We formalize \textit{constitution learning} and propose Multi-Agent Constitutional Learning (MAC), where specialized agents propose, critique, and refine rules.
\item We evaluate MAC on PII tagging across finance, legal, and healthcare domains, demonstrating over 50\% improvement over prompt optimization baselines and performance comparable to supervised fine-tuning.
\item We introduce two variants: reMAC, which uses test-time retrieval, and MAC+, which trains agents on successful trajectories to reinforce effective updates.
\item We show that MAC generalizes beyond classification to agentic tasks by adapting it to tool calling on BFCL.
\end{itemize}

\begin{table*}[t]
\centering
\small
\caption{Example natural-language constitutional rules learnt by MAC for private information tagging across different domains.}
\label{tab:constitution_rules}
\begin{tabular}{@{}p{1.5cm} p{14.5cm}@{}}
\toprule
\textbf{Domain} & \textbf{Example Constitutional Rules} \\
\midrule
\textbf{Legal} &
Mark as private specific dates when they appear in the context of personal events or actions, such as births, deaths, or significant life events. Do not mark general references or narrative text containing similar time periods. 
\emph{Examples:} mark ``1975'', ``22 August 2003''; do not mark ``on a day in June''. \\[0.4em]
\bottomrule
\textbf{Healthcare} &
Mark terms such as heart failure subtypes (e.g., diastolic heart failure, systolic heart failure) when explicitly mentioned in a patient's medical history as private. Do not mark generic medical conditions without an explicit subtype or detailed description. \\[0.4em]
\bottomrule
\textbf{Finance} &
Mark as private any phrase indicating a specific financial timeframe (e.g., FY2022, YTD FY2021) when it appears in direct association with identifiable information. Do not mark generic narrative context or standalone labels without specific identifiers. 
\emph{Examples:} mark ``FY2022 balance sheet'', ``Q4 2021 sales report'' (if associated with identifiable information); do not mark ``next year'', ``past fiscal year'' without identifiers. \\
\bottomrule
\end{tabular}
\end{table*}

\vspace{-20pt}
\section{Background}

\paragraph{Controlling LLM Behavior.}
Large language models can be controlled through two approaches: training-time interventions and inference-time interventions.
Training-time methods include supervised fine-tuning (SFT) and reinforcement learning from human feedback (RLHF).
These methods are effective but require vast labeled data, computational resources, and risk memorization of sensitive training examples \cite{carlini2023quantifyingmemorizationneurallanguage}.
Inference-time methods instead guide behavior through the prompt, leaving model weights unchanged and includes few-shot prompting~\cite{brown2020language}, chain-of-thought prompting~\cite{wei2022chain}, and rule-based prompting~\cite{ouyang2022training}.

\textbf{Constitutional AI} guides LLM behavior at inference time using an explicit constitution \cite{bai2022constitutional,huang2024collective}: a set of natural-language rules, as illustrated in Table~\ref{tab:constitution_rules}. 
By placing constraints on the prompt's structure, it improves transparent and auditable control. 
For example, constitutional classifiers apply such rules to judge whether input text is harmful or non-harmful \cite{sharma2025constitutional}.

\begin{figure}
    \centering
    \includegraphics[width=0.98\linewidth]{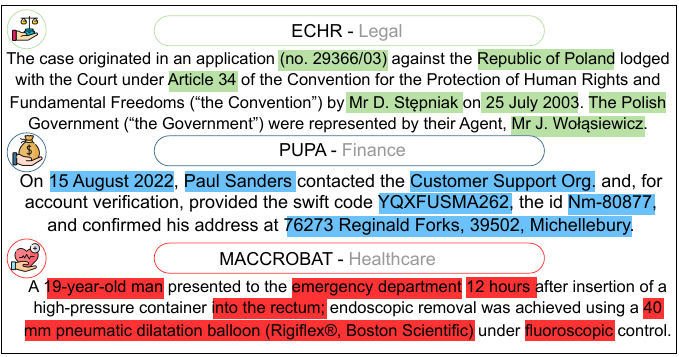}
    \caption{Example document snippets with private spans indicated.}
    \label{fig:maccrobat_example}
\end{figure}

\textbf{Automatic Prompt Optimization (APO)} uses LLMs to iteratively refine prompts based on task performance~\cite{ramnath2025systematic}.
Given labeled examples and a task metric, APO methods generate candidate modifications, evaluate them, and retain improvements.

We survey evolutionary and proposal-based prompt optimization methods, such as Genetic-Pareto (GEPA) \cite{GEPA}, which performs guided evolutionary search over prompts using task feedback, and Multiprompt Instruction Proposal Optimizer (MIPRO) \cite{MIPRO}, which iteratively proposes and selects prompts based on validation performance. 
Both methods optimize prompts as unstructured text, putting no constraints on their format. 
%


\textbf{Private Information Tagging} aims to identify sensitive text spans in documents.
These spans may correspond to direct identifiers such as names or account numbers, or indirect references that become sensitive\footnote{Definitions of what constitutes private information vary across organizations, jurisdictions, and datasets, and can evolve over time.} in context \cite{PII_EXP_1, PII_EXP_2}.
Existing approaches typically rely on pretrained taggers, rule-based pattern matching, or context-aware detectors operating over fixed label sets \cite{microsoft_presidio_docs, nvidia_glinerPII_2025, gliner}.

\textbf{Notation.}
Let $x \in \mathcal{X}$ denote an input (e.g., a document) and $y^\star(x) \in \mathcal{Y}$ denote the ground-truth output (e.g., a set of labeled spans), which reflects a task-specific definition revealed only through labeled examples. 
The agent is guided by a constitution $C = \langle r_1, \ldots, r_{|C|} \rangle$, which is an ordered list of natural-language rules. 
Let $\theta$ denote fixed LLM weights and $\pi$ a fixed role prompt specifying the task and output format. 
Given an input $x$ and constitution $C$, the agent produces a prediction $\hat{y}(x) = A_{\theta, \pi}(x; C)$. 
We assume access to a labeled dataset $D = \{(x_i, y_i^\star)\}_{i=1}^N$ and a verifiable task metric $\mu : \mathcal{Y} \times \mathcal{Y} \rightarrow [0,1]$ that compares predictions to ground truth.

\section{Conceptual Approach}
We begin by describing our core algorithm for MAC and then we describe two extension variants MAC+ and reMAC. 

\subsection{Constitutional Learning}
The goal of constitution learning is to automatically derive a set of interpretable rules from labeled examples. 
Given training data $D_{\text{train}}$ and validation data $D_{\text{val}}$, and a task metric $\mu : \mathcal{Y} \times \mathcal{Y} \rightarrow [0,1]$, we seek a constitution $C$ that maximizes task performance.
Formally, we optimize:
\begin{align}
C^\star = \arg\max_{C \in \mathcal{C}} \ \mu\left( A_{\theta, \pi}(D_{\text{val};  C}), Y^\star_{\text{val}}  \right)
\end{align}
where $A_{\theta, \pi}(x; C)$ denotes the prediction of an LLM-based agent with fixed weights $\theta$ and role prompt $\pi$, conditioned on input $x$ and constitution $C$.
Throughout optimization, the model weights $\theta$ and role prompt $\pi$ remain fixed.

\subsection{Multi-Agent Constitution Learning (MAC)}
We call our algorithm \emph{Multi-Agent Constitution Learning} (MAC), which optimizes constitutions through a network of specialized agents.
MAC consists of four agents:
\begin{itemize}[itemsep=0mm]
    \item \textsc{Annotator}($B$, $C$): applies the current constitution $C$ to a batch of inputs $B$ and produces predictions $\hat{Y}$.
    \item \textsc{Decision}($FN$, $FP$, $C$, $R_{\max}$): analyzes errors and selects an update operation $op \in \{\textsc{Add}  \textsc{Edit}, \textsc{Remove}\}$, along with a rationale $r$ and target rule index $i$.
    \item \textsc{Creator}($FN$, $FP$, $r$): drafts a new rule based on observed errors and the rationale.
    \item \textsc{Editor}($c_i$, $FN$, $FP$, $r$): modifies existing rule $c_i$ based on observed errors and the rationale.
\end{itemize}
Given predictions $\hat{Y}$ and ground-truth labels $Y^\star$, a helper function \textsc{Match} computes the task metric $\mu(\hat{Y}, Y^\star)$ and extracts false negatives ($FN$) and false positives ($FP$), along with their surrounding context within a fixed character window. Because private information tagging is a span-level classification task, we use F1 as the standard evaluation metric, computed (using \textsc{Match}) via exact span-level matching between predicted and ground-truth private phrases. 


%
Using multiple agents allows decomposing the optimization problem into subtasks, so that each agent can focus on its own individual task, which has been shown to improve reliability~\cite{chan2023chateval, chen2024reconcile, du2023improving}.
For reproducibility\footnote{We will also publish our software with all details to reproduce our experiments, including the learned constitutions as open-source code.}, we provide prompts for the Annotator, Decision, Rule Creator, and Rule Editor agents in the Appendices~\ref{app:annotator}, \ref{app:decision}, \ref{app:creator}, and \ref{app:editor}. 
Appendix~\ref{app:sample_outputs_mac} presents sample outputs from individual MAC agents.

\subsection{Core Algorithm for MAC}
MAC optimizes the constitution through an iterative epoch-batch procedure, summarized in Algorithm~\ref{alg:constitution_classifier}.
At each step, the \textsc{Annotator} applies the current constitution $C$ to a batch $B$ (Line 5).
The \textsc{Match} operation computes errors and the task metric $\mu$ (Line 6).
The \textsc{Decision} agent analyzes the errors and selects an update operation (Line 7).
Based on this decision, either the \textsc{Creator} or \textsc{Editor} agent proposes an updated constitution $C_{\text{prop}}$ (Lines 8--14).

The proposed constitution is evaluated on the same batch (Line 15).
If the task metric improves ($\mu_{\text{prop}} > \mu$), the update is accepted (Lines~16–18). Otherwise, the proposal is retried up to a fixed number of times using a higher sampling temperature; if no retry yields an improvement, the original constitution is retained.
After a fixed number of batches, the constitution is evaluated on the validation set $D_{\text{val}}$, and the version with the highest validation performance is retained as $C^\star$ (Lines 20--23). Validation F1 over batches for MAC runs across different models and datasets is shown in Appendix~\ref{app:constitutional_evolution}.

\begin{algorithm}[tb]
\small
\caption{\textsc{MAC}: Multi-Agent Constitution Learning}
\label{alg:constitution_classifier}
\begin{algorithmic}[1]
\STATE {\bfseries Input:} Training data $D_{\text{train}}$, validation data $D_{\text{val}}$, rule budget $R_{\max}$, epochs $E$
\STATE {\bfseries Initialize:} $C \leftarrow \varnothing$, $C^\star \leftarrow \varnothing$, $\mu^\star \leftarrow 0$

\FOR{$e = 1$ {\bfseries to} $E$}
  \FORALL{batch $B \subset D_{\text{train}}$}
    \STATE $\hat{Y} \leftarrow \textsc{Annotator}(B, C)$
    \COMMENT{Apply current constitution}
    \STATE $FN, FP, \mu \leftarrow \textsc{Match}(\hat{Y}, Y^\star_B)$
    \COMMENT{Compute errors}
    \STATE $op, r, i \leftarrow \textsc{Decision}(FN, FP, C, R_{\max}$)
    \IF{$op = \textsc{Add}$}
      \STATE $C_{\text{prop}} \leftarrow C \cup \{\textsc{Creator}(FN, FP, r)\}$
    \ELSIF{$op = \textsc{Edit}$}
      \STATE $C_{\text{prop}} \leftarrow (C \setminus \{c_i\}) \cup \{\textsc{Editor}(c_i, FN, FP, r)\}$
    \ELSE
      \STATE $C_{\text{prop}} \leftarrow C \setminus \{c_i\}$
      \COMMENT{\textsc{Remove}}
    \ENDIF
    \STATE $\mu_{\text{prop}} \leftarrow \textsc{Match}(\textsc{Annotator}(B, C_{\text{prop}}), Y^\star_B)$
    \IF{$\mu_{\text{prop}} > \mu$}
      \STATE $C \leftarrow C_{\text{prop}}$
      \COMMENT{Accept update}
    \ENDIF
  \ENDFOR
  \STATE $\mu_{\text{val}} \leftarrow \textsc{Match}(\textsc{Annotator}(D_{\text{val}}, C), Y^\star_{\text{val}})$
  \IF{$\mu_{\text{val}} > \mu^\star$}
    \STATE $C^\star \leftarrow C$, $\mu^\star \leftarrow \mu_{\text{val}}$
    \COMMENT{Track best}
  \ENDIF
\ENDFOR

\STATE {\bfseries Return:} $C^\star$
\end{algorithmic}
\end{algorithm}

\subsection{Two Variants: MAC+ and reMAC}
\paragraph{MAC+.} The effectiveness of MAC depends on each agent performing its role reliably, but smaller models (ex. 3B) may struggle with complex reasoning tasks such as diagnosing error patterns or proposing precise rules.
MAC+ addresses this by fine-tuning each agent on high-quality trajectories generated by a larger model (ex. 14B).
Let $A^{(S)}$ denote an agent instantiated with a smaller model and $A^{(L)}$ the same agent with a larger model.
We execute MAC using $A^{(L)}$ and collect input-output traces $\mathcal{T}_A = \{(z_j, o_j)\}_{j=1}^{M_A}$ for each agent $A \in \{\textsc{Decision}, \textsc{Creator}, \textsc{Editor}\}$, where $z_j$ is the input context and $o_j$ is the output produced by $A^{(L)}$.
We then fine-tune $A^{(S)}$ on $\mathcal{T}_A$ via supervised fine-tuning by training each agent only on traces relevant to its role.
To avoid memorization, we fine-tune on data from different domains (\textit{healthcare \& finance}) than those used for evaluation (\textit{legal}), training agents on task structure rather than dataset-specific content. Statistics on the collected trajectories from each agent used for fine-tuning are shown in Appendix~\ref{app:rollout_statistics}.

\begin{figure}[tb]
    \centering
    \includegraphics[width=\columnwidth]{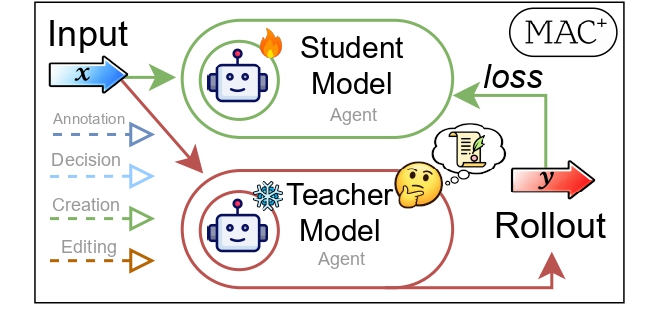}    
    \caption{\textsc{MAC+}: Optimizing each agent for its assigned role via supervised fine-tuning on rollouts generated by a larger model.}
    \label{fig:mac_plus}
\end{figure}

\begin{figure}[tb]
    \centering
    \includegraphics[width=0.9\columnwidth]{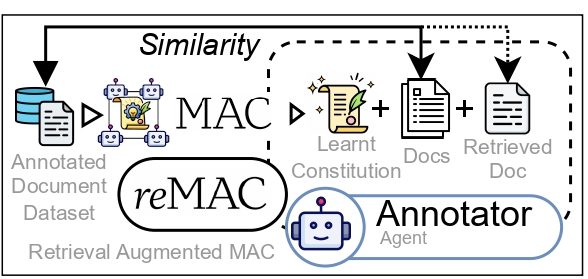}    
    \caption{reMAC: Retrieval-augmented MAC, where documents from the training set are retrieved and added at test time.}
    \label{fig:remac}
\end{figure}

\paragraph{Retrieval-Augmented MAC (reMAC).}
MAC learns a constitution that captures general domain patterns, but abstract rules may be insufficient for highly specialized terminology or context-dependent definitions.
reMAC mitigates this by supplementing the learned constitution with concrete examples at test time.
Let $\mathcal{R}: \mathcal{X} \times D_{\text{train}} \times k \rightarrow \mathcal{X}^k$ be a retrieval function returning $k$ training examples for a test input $x$.
The \textsc{Annotator} prediction becomes:
\begin{align}
\hat{y}(x) = A_{\theta, \pi}\big(x; C, \mathcal{R}(x, D_{\text{train}}, k)\big)
\end{align}
where retrieved examples are included as few-shot demonstrations alongside the constitution $C$.
Importantly, as shown in \Cref{fig:remac} retrieval is applied only at inference and does not modify the constitution so that its auditability is preserved.

\section{Evaluation}
We evaluate MAC on private information tagging across legal, healthcare, and finance domains and then show generalization to other tasks, such as tool calling.
We focus on tagging because it combines the core challenges constitution learning addresses: (i) classification, (ii) limited labeled data, (iii) a need for transparent, auditable decisions, and (iv) settings where concerns about memorization make training-based approaches undesirable.

\subsection{Experimental Setup}
\paragraph{Datasets.}
We evaluate MAC on private information tagging across three domains: legal, healthcare, and finance.
For legal, we use the ECHR dataset \cite{echr_dataset}, containing case documents from the European Court of Human Rights.
For healthcare, we use MACCROBAT \cite{maccrobat2020}, containing clinical documents with dense, overlapping private spans across 49 fine-grained privacy types.
For finance, we use a subset of PUPA \cite{li2025papillon}, derived from real-world user-assistant interactions.
These domains differ substantially in privacy definitions, document structure, and annotation density, enabling evaluation across sparse-privacy (legal), heavy-redaction (healthcare), and intermediate (finance) regimes.
We sample 192 documents per domain, split into 160 train, 16 validation, and 16 test documents. We use 192 documents because the smallest dataset (MACCROBAT) contains 192 usable documents, and we apply the same split consistently across all domains for consistency.
Full statistics are provided in Appendix~\ref{app:dataset_stats}.

\paragraph{Models and Metrics.}
We use Qwen~2.5 Instruct models \cite{qwen2025qwen25technicalreport} at 3B, 7B, and 14B parameters, as they are standard non-reasoning instruction-tuned models with a complete size family and are commonly used in prior prompt-optimization work. We report F1 computed via exact span-level matching between predicted and ground-truth private spans, which is the standard evaluation metric for this task \cite{sang2003introduction}.


\paragraph{Baselines.}
We compare MAC against 3 types of baselines:
(i) \textit{Pretrained taggers}: Microsoft Presidio \cite{microsoft_presidio_docs} with SpaCy, BERT, and Stanza backends, and Nvidia GLiNER-PII \cite{nvidia_glinerPII_2025}.
(ii) \textit{Prompt optimization}: GEPA \cite{GEPA}, which performs evolutionary prompt search, and MIPRO \cite{MIPRO}, which iteratively proposes and selects prompt variants. Both are configured with the same computational budget as MAC (480 forward passes).
(iii) \textit{Training-based}: Supervised fine-tuning (SFT) and GRPO \cite{GRPO}, included as upper-bound references despite memorization concerns.
Full implementation details are available in Appendix~\ref{app:baseline_details}.

\begin{table}
\centering
\small
\caption{Performance comparison (F1) with pretrained taggers.}
\label{tab:pretrained_comparison}
\begin{tabular}{@{}lrrr@{}}
\toprule
Method & Legal & Finance & Healthcare \\
\midrule
MAC (14B) & \textbf{67.3} & \textbf{45.5} & \textbf{26.7} \\
\midrule
Microsoft Presidio (SpaCy)  & 57.3 & 32.8 & 12.0 \\
Microsoft Presidio (BERT)   & 10.4 &  5.2 &  2.7 \\
Microsoft Presidio (Stanza) & 54.0 & 44.7 &  9.1 \\
Nvidia GLiNER     & 57.1 & 35.9 &  5.3 \\
\bottomrule
\end{tabular}
\end{table}

\paragraph{Computational Budget.}
For fair comparison, all optimization-based methods use the same computational budget.
MAC runs for 3 epochs over 20 batches of 8 documents, totaling 480 forward passes.
GEPA and MIPRO are configured with equivalent budgets: GEPA uses $\texttt{max\_metric\_calls}=480$ and MIPRO is capped at 480 total evaluations.
Training-based baselines use the same data splits and epoch count.
Full implementation details are provided in Appendix~\ref{app:baseline_details}. 

\subsection{Experimental Results}
\begin{figure*}
    \centering
    \includegraphics[width=1.0\textwidth,height=4cm]{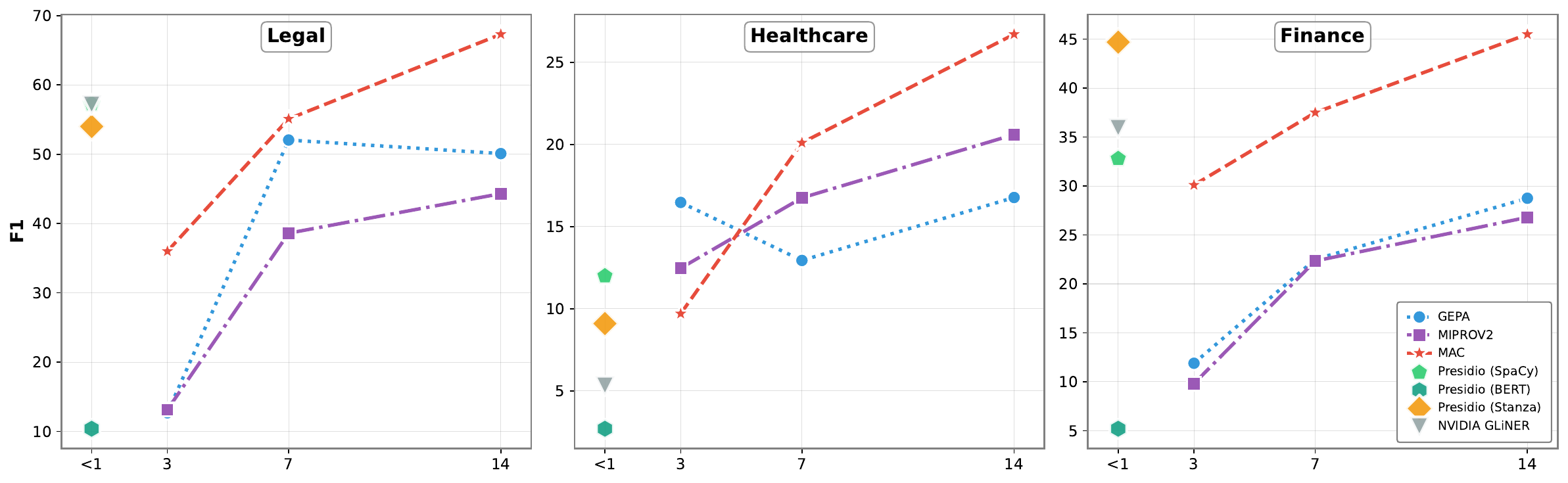}
    \caption{Test F1 across model sizes for MAC compared to baseline prompt optimizers and pretrained PII taggers.}
    \label{fig:overall_combined}
\end{figure*}

Figure~\ref{fig:overall_combined} presents a comparison of test F1 between MAC, pretrained taggers, and prompt-optimization baselines (GEPA \& MIPRO) across model sizes (3B, 7B, \& 14B).

\paragraph{MAC matches or outperforms pretrained taggers.}
Table~\ref{tab:pretrained_comparison} compares MAC (14B) against pretrained PII taggers commonly used in practice, including Microsoft Presidio with different NER backends and Nvidia GLiNER-PII.

MAC outperforms all baselines across domains: +17\% in Legal (67.3 vs.\ 57.3), +2\% in Finance (45.5 vs.\ 44.7), and +123\% in Healthcare (26.7 vs.\ 12.0).
The gain is largest in Healthcare, where pretrained taggers struggle with fine-grained medical privacy types that fall outside standard PII categories.
In Finance, where privacy definitions more closely align with standard PII, the gap is smaller but MAC still achieves the best performance.
Unlike pretrained taggers, which rely on fixed label sets, MAC adapts to domain-specific privacy definitions through the constitutions alone.

\begin{table}
\centering
\small
\caption{Performance comparison (F1) across model sizes.}
\label{tab:optimizer_modelsizes_compact}
\begin{tabular}{@{}llrrr@{}}
\toprule
Dataset & Optimizer & 3B & 7B & 14B \\
\midrule
\multirow{3}{*}{Legal} 
 & GEPA       & 12.70 & 52.05 & 50.09 \\
 & MIPRO    & 13.15 & 38.61 & 44.31 \\
 & MAC        & \textbf{36.00} & \textbf{55.10} & \textbf{67.30} \\
\midrule
\multirow{3}{*}{Finance}
 & GEPA       & 11.89 & 22.51 & 28.75 \\
 & MIPRO    &  9.81 & 22.34 & 26.80 \\
 & MAC        & \textbf{30.10} & \textbf{37.50} & \textbf{45.50} \\
\midrule
\multirow{3}{*}{Healthcare}
 & GEPA       & \textbf{16.47} & 12.94 & 16.77 \\
 & MIPRO    & 12.47 & 16.75 & 20.60 \\
 & MAC        &  9.70 & \textbf{20.10} & \textbf{26.70} \\
\bottomrule
\end{tabular}
\end{table}

\paragraph{MAC outperforms prompt optimization baselines.}
Table~\ref{tab:optimizer_modelsizes_compact} compares MAC against GEPA and MIPRO across model sizes.
MAC outperforms both baselines in 8 of 9 dataset-model configurations.
The largest gain is observed in Legal at 3B (+174\%, 36.0 vs.\ 13.15 F1).
The only case where MAC underperforms is Healthcare at 3B, where GEPA achieves 16.47 vs.\ 9.70 F1.
In all other settings, improvements range from +28\% to +153\%.
These results demonstrate that structured rule optimization consistently outperforms unstructured prompt search.

\begin{table}[t]
\centering
\small
\caption{reMAC: Retrieval-augmented MAC. F1 for varying numbers of retrieved examples added to MAC constitutions.}
\label{tab:remac_performance}
\begin{tabular}{@{}llrrrr@{}}
\toprule
\textbf{Dataset} & \textbf{Model} & \textbf{0-Shot} & \textbf{1-shot} & \textbf{3-shot} & \textbf{5-shot} \\
\midrule
\multicolumn{6}{@{}l}{\textit{Random Retrieval:}} \\
\midrule
\multirow{3}{*}{Legal} 
          & 3B  & 38.79 & 40.94 & \textbf{45.64} & 33.17 \\
          & 7B  & 54.85 & \textbf{64.93} & 62.29 & 60.56 \\
          & 14B & 73.41 & 77.35 & \textbf{77.74} & 76.29 \\
\cmidrule(lr){1-6}
\multirow{3}{*}{Finance} 
          & 3B  & 17.42 & 21.82 & \textbf{25.18} & 25.06 \\
          & 7B  & 38.79 & 32.32 & 35.88 & \textbf{46.73} \\
          & 14B & 48.47 & \textbf{49.55} & 40.60 & 42.61 \\
\cmidrule(lr){1-6}
\multirow{3}{*}{Healthcare} 
          & 3B  & 16.43 & 31.72 & 24.99 & \textbf{34.35} \\
          & 7B  & 10.06 & 22.84 & \textbf{28.43} & 25.27 \\
          & 14B & 16.16 & 34.18 & 46.51 & \textbf{47.00} \\
\midrule
\multicolumn{6}{@{}l}{\textit{Similarity-Based Retrieval:}} \\
\midrule
\multirow{3}{*}{Legal} 
          & 3B  & 42.56 & 41.02 & 52.59 & \textbf{58.15} \\
          & 7B  & 56.36 & \textbf{66.36} & 63.66 & 59.30 \\
          & 14B & 72.21 & 78.48 & 79.39 & \textbf{79.97} \\
\cmidrule(lr){1-6}
\multirow{3}{*}{Finance} 
          & 3B  & 13.82 & 25.95 & \textbf{33.92} & 22.39 \\
          & 7B  & 38.79 & 47.88 & 45.33 & \textbf{50.60} \\
          & 14B & 48.47 & 46.64 & 46.19 & \textbf{49.95} \\
\cmidrule(lr){1-6}
\multirow{3}{*}{Healthcare} 
          & 3B  & 16.97 & 32.99 & \textbf{33.88} & 32.78 \\
          & 7B  &  9.91 & 28.85 & 36.63 & \textbf{36.97} \\
          & 14B & 16.34 & 39.63 & 46.36 & \textbf{49.60} \\
\bottomrule
\end{tabular}
\end{table}

\paragraph{reMAC improves utility during inference.}
Table~\ref{tab:remac_performance} shows results for reMAC, which augments the learned constitution with retrieved examples at test time. While we evaluate $k \in {1,2,3,4,5}$, we report the representative settings $k \in {1,3,5}$ here for brevity; Appendix~\ref{app:remac_full_plots} presents the full test F1 curves over $k$ for all models and datasets.
We evaluate two retrieval strategies: random sampling from the training set, and similarity-based retrieval using cosine similarity over sentence embeddings\footnote{We use the following pre-trained sentence transformer: \href{https://huggingface.co/sentence-transformers/all-MiniLM-L6-v2}{https://huggingface.co/sentence-transformers/all-MiniLM-L6-v2}}.

Retrieval consistently improves performance across all 18 dataset-model configurations.
The largest gain is observed in Healthcare at 7B with similarity-based retrieval (+273\%, 9.91 to 36.97 F1 at 5-shot).
Averaged across settings, reMAC yields improvements of 75.5\% (3B), 90.4\% (7B), and 69.4\% (14B).
Gains are largest in Healthcare (176.5\%), where specialized medical terminology benefits most from concrete examples.
These results show that test-time retrieval complements learned constitutions, particularly for domains with context-dependent definitions.
\begin{table}
\centering
\small
\caption{Optimized MAC vs. training-based baselines on legal for the 3B Model. MAC+ tunes on Healthcare \& Finance only.}
\label{tab:3b_baseline_comparison}
\begin{tabular}{@{}lrrr@{}}
\toprule
\textbf{Method} & \textbf{Precision} & \textbf{Recall} & \textbf{F1} \\
\midrule
\multicolumn{4}{@{}l}{\textit{Training Baselines:}} \\
\midrule
\addlinespace[0.2em]
GRPO & 64.61 & 28.63 & 39.68 \\
SFT  & \textbf{78.72} & 44.17 & 56.59 \\
\addlinespace[0.5em]
\midrule
\multicolumn{4}{@{}l}{\textit{MAC Variants:}} \\
\midrule
\addlinespace[0.2em]
MAC  & 64.32 & 24.95 & 35.96 \\
MAC+ & 75.93 & \textbf{48.27} & \textbf{59.02} \\
\bottomrule
\end{tabular}
\end{table}

\paragraph{MAC+ improves over MAC.}
Table~\ref{tab:3b_baseline_comparison} compares MAC+ against training-based baselines on ECHR (Legal) using a 3B model.
MAC+ fine-tunes each agent on trajectories generated by a 14B model executing MAC on out-of-domain data (healthcare and finance), then evaluates on legal documents.
Loss curves for all agents are shown in Appendix~\ref{app:mac_plus_loss_curves}, and we fine-tune agents using LoRA (rank 64). 

\begin{figure}
    \centering
    \includegraphics[width=0.98\columnwidth]{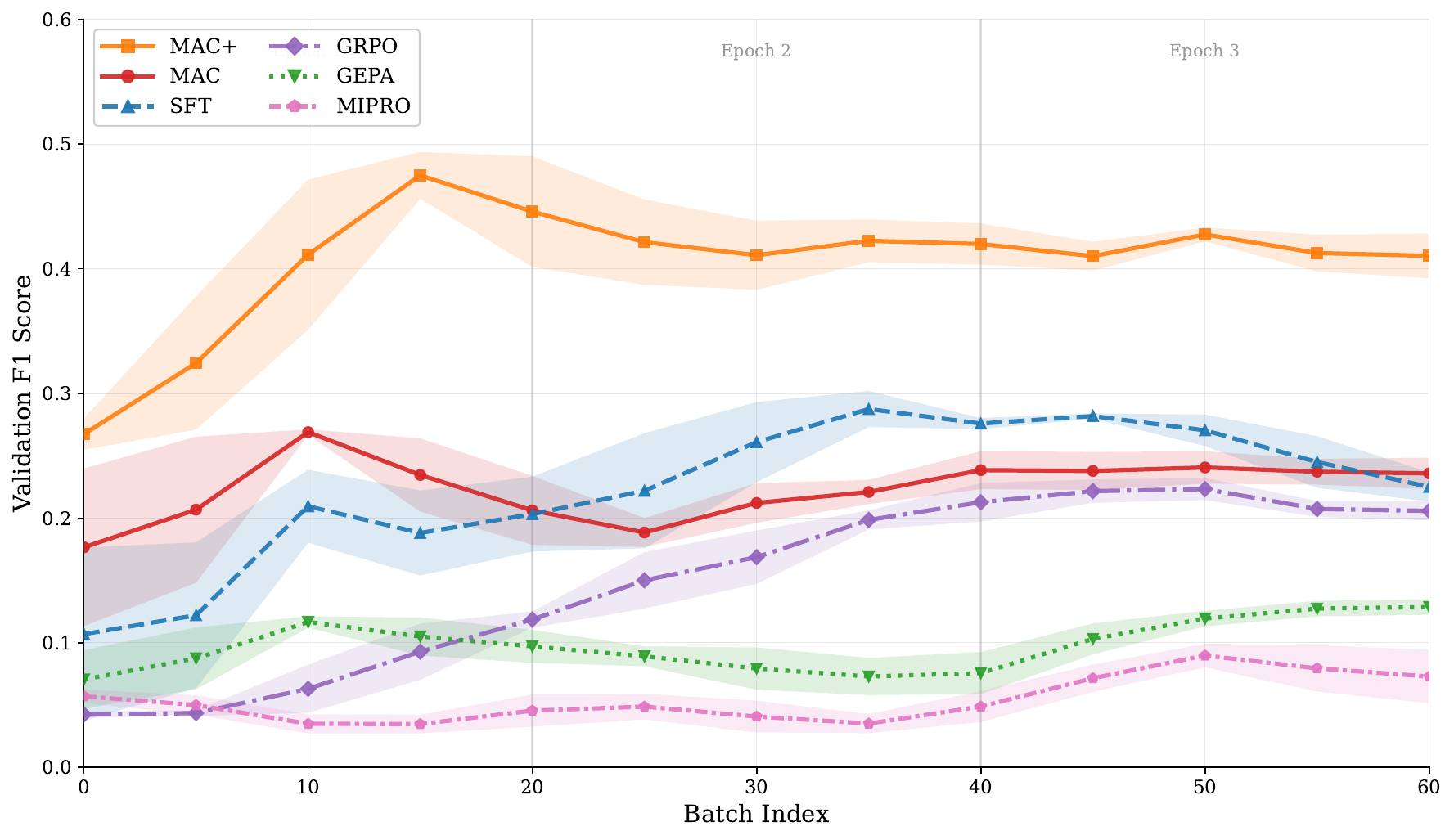}
    \caption{Validation F1 vs training batches comparison of MAC variants against baselines on 3B models for the ECHR dataset.}
    \label{fig:3b_comparison}
\end{figure}

This setup trains agents on task structure rather than domain-specific patterns, avoiding memorization.
MAC+ achieves 59.02 F1, outperforming both SFT (56.59) and GRPO (39.68).
Gains are primarily from higher recall (48.27 vs.\ 44.17 for SFT) while maintaining comparable precision (75.93 vs.\ 78.72), indicating that specialized agents identify more private spans without sacrificing accuracy.
Figure~\ref{fig:3b_comparison} shows that MAC+ also converges faster, exceeding 0.4 F1 within approximately 10 batches while no baseline reaches this threshold.
These results demonstrate that agent-level distillation improves performance for smaller models without requiring in-domain training data.
%


\begin{table}[t]
  \centering
  \small
  \caption{Performance of MAC+ agents. The Annotator is evaluated using F1 on the Legal test set. All other agents are evaluated using an LLM-as-a-judge win rates (remaining cases are ties).}
  \label{tab:agent_lora_comparison}
  \begin{tabular}{lcc}
  \toprule
  \textbf{Agent} & \textbf{Base Model} & \textbf{MAC+} \\
  \midrule
  \multicolumn{3}{l}{\textit{Classification (F1 score)}} \\
  \addlinespace[0.5ex]
  Annotator Agent       & 32.9  & 59.02 \\
  \cmidrule(lr){1-3}
  \addlinespace[0.5ex]
  \multicolumn{3}{l}{\textit{LLM-as-a-Judge (Win rate)}} \\
  \addlinespace[0.5ex]
  Decision Agent        & 9.0\%  & 86.0\% \\
  Rule Editor Agent     & 25.0\% & 74.0\% \\
  Rule Proposer Agent   & 3.0\%  & 97.0\% \\
  \bottomrule
  \end{tabular}
\end{table}

\paragraph{Analysis of MAC+.}
To understand the source of MAC+ improvements, we evaluate each agent individually.
We construct an evaluation set of 100 input-output traces per agent. For the Annotator, we use trajectories collected from MAC executions with the 14B model (for which we have sufficient samples). For the remaining agents, we supplement MAC-collected trajectories with additional samples generated by a 7B model.
For each trace, we re-run the input through both the base model and the fine-tuned model, enabling direct comparison under identical inputs.
The Annotator is evaluated using F1. Other agents are evaluated using an LLM-as-a-judge protocol with GPT-OSS 120B (prompt in Appendix~\ref{app:llm_as_a_judge}).

Table~\ref{tab:agent_lora_comparison} shows substantial improvements across all agents after using MAC+. 
The Annotator improves from 32.9 to 59.02 F1 (1.8$\times$). Win rates increase from 9.0\% to 86.0\% for Decision (9.6$\times$), from 25.0\% to 74.0\% for Rule Editor (3.0$\times$), and from 3.0\% to 97.0\% for Rule Proposer (32$\times$).
These results confirm that MAC+ gains stem from improved performance on each agent's specialized sub-task, rather than from any single component.

\paragraph{Tool-MAC: MAC for agentic tool-calling.}
To demonstrate that MAC applies beyond private information tagging, we evaluate on the BFCL tool-calling benchmark \cite{patil2025bfcl}, specifically the \emph{parallel} sub-task that requires decomposing a single natural-language request into multiple independent function calls.
We choose tool calling because it shares key properties with PII tagging: both require precise structured outputs, and tool arguments can contain sensitive information that must be handled correctly \cite{alizadeh2025simple, wang2025unveiling, yan2025protecting}.

To adapt MAC for tool calling, we modify the agent prompts to reason about function signatures rather than privacy definitions, replace the optimization metric (accuracy instead of F1), and adapt the \textsc{Match} operation to identify failed tool calls rather than false positives or negatives.
Table~\ref{tab:tool_calling_comparison} shows results for a 3B model.
MAC improves accuracy by +5.5\% over the 82\% baseline, outperforming GEPA (+4.0\%) and MIPRO (+2.5\%).
The learned rules address common failure modes, such as mismatched argument types, and improve tool selection for multi-step user intents.
These results show that MAC can learn constitutions for diverse downstream tasks, with adaptations limited to task-specific prompts and metrics while the core multi-agent optimization loop remains unchanged.

\subsection{Qualitative Analysis}

The first three rules from all constitutions learned by MAC across Legal, Healthcare, and Finance datasets and across the 3B, 7B, and 14B model sizes are provided in Appendix~\ref{app:mac_constitutions}. GEPA and MIPRO learned prompts for the Legal dataset across the same three model sizes are provided in Appendices~\ref{app:gepa_prompts} and \ref{app:mipro_prompts}, respectively.

On average, MAC learns 9 rules across datasets and model sizes, following a consistent structure similar to the sample rules in Table~\ref{tab:constitution_rules}. Rules are typically phrased as “Mark …” definitions of private entities, optionally augmented with examples or explicit inclusion/exclusion criteria, ensuring transparency and enabling direct stakeholder inspection and updates. Legal and Healthcare require a comparable number of rules (7.33 and 7.67), while Finance requires more on average (12), which we attribute to its sparser, shorter private entities and lowest average characters per privacy group (Appendix~\ref{app:dataset_stats}), necessitating more independent rules to precisely guide span boundaries and improve performance. Overall, the length of full MAC prompts (with learned constitutions) lies between MIPRO and GEPA (Table~\ref{tab:method_averages}). On closer inspection of the learned constitutions, we observe that smaller models (3B) exhibit recurring artifacts, including overly short or low-quality rules with substantial repetition, incorrect edit decisions, and redundant updates, with the Decision Agent showing limited diversity and a bias toward adding new rules. These issues are substantially reduced when using larger models (14B), motivating the use of distillation in MAC+, which yields significant performance improvements. Despite these limitations, MAC still outperforms GEPA for smaller models. GEPA’s reflection and full prompt rewrites make it difficult for low-capacity models to preserve task-relevant behavior, whereas MAC constrains updates to the constitution and decomposes the problem across specialized agents, enabling the discovery of performance-improving rule sets even with a 3B model.

\begin{table}
\centering
\small
\caption{Tool-calling accuracy after optimization. Baseline = 82\%.}
\label{tab:tool_calling_comparison}
\begin{tabular}{lcc}
\toprule
Method & Final Accuracy & Improvement \\
\midrule
MAC     & 87.5\% & +5.5\% \\
GEPA    & 86.0\% & +4.0\% \\
MIPRO   & 84.5\% & +2.5\% \\
\bottomrule
\end{tabular}
\end{table}


\section{Discussion \& Limitations}

\textbf{MAC is particularly effective for smaller models.}
Table~\ref{tab:optimizer_modelsizes_compact} shows that the performance gap between MAC and prompt-optimization baselines decreases as model size increases. The average improvement over GEPA drops from 98.5\% at 3B to 50.6\% at 14B. We attribute this to how MAC structures the optimization problem. Rather than rewriting the full prompt, MAC restricts updates to explicit rules while keeping the rest of the prompt fixed. MAC further decomposes updates into decision, rule creation, and rule editing stages handled by separate agents, limiting the complexity of each LLM call. This decomposition is particularly beneficial for smaller models, which struggle with complex multi-step reasoning in a single call.

\textbf{When does MAC underperform?}
MAC underperforms GEPA in one setting: Healthcare at 3B (9.70 vs. 16.47 F1). We hypothesize this occurs because healthcare privacy definitions in MACCROBAT span 49 fine-grained types with dense, overlapping spans. At 3B, the \textsc{Decision} agent may lack the capacity to diagnose such complex error patterns, leading to suboptimal rule updates. This suggests MAC may struggle when (i) the task requires fine-grained distinctions beyond the agent's reasoning capacity, or (ii) error patterns are too complex to address with single-rule updates. Future work could explore hierarchical rule structures or ensemble decision-making to address such cases.

\textbf{Constitutional prompts are preferred for transparency.}
As model capacity increases, the performance gap between MAC and baselines narrows. However, MAC remains preferable from a transparency perspective. MAC learns structured rules that follow a fixed format (Table~\ref{tab:constitution_rules}), enabling auditing beyond raw performance. Example constitutions are provided in Appendix~\ref{app:mac_constitutions}, with GEPA and MIPRO prompts in Appendices~\ref{app:gepa_prompts} and \ref{app:mipro_prompts}. GEPA prompts are substantially more verbose than MAC and MIPRO (Table~\ref{tab:method_averages}). On average, MAC prompts are approximately 30\% shorter than GEPA prompts, and up to 70\% shorter on Legal data. Because the base prompt remains fixed and only the constitution evolves, stakeholders do not need to repeatedly inspect entirely new prompts for each dataset or task.



\textbf{Agent specialization without memorization.}
MAC+ trains agents on out-of-domain data (healthcare and finance) and evaluates on legal documents, learning task structure rather than domain-specific patterns. This is critical for private information tagging, where fine-tuning on sensitive data risks memorization \cite{carlini2023quantifyingmemorizationneurallanguage}. MAC+ outperforms both SFT and GRPO on out-of-domain evaluation (Table~\ref{tab:3b_baseline_comparison}), suggesting that agent-level distillation provides a safer alternative to direct fine-tuning when privacy is a concern.

\textbf{Limitations.}
While we perform extensive evaluations of our methods, we identify the following limitations. 
First, the length of the constitution presents a utility/transparency trade-off where adding rules improves utility but makes them more difficult to audit. 
We do not evaluate this trade-off, which would require a user-study and likely exceeds the scope of our work. 
Second, our experiments assumes the task metric is reliable. 
However, if $\mu$ is noisy, sparse, or misaligned with the true objective, MAC may learn rules that optimize the metric without improving actual task performance. 
Third, rules can interact in unexpected ways when the constitution grows too large.
Adding a new rule may conflict with existing rules or cause unintended side effects on previously correct predictions. 
This may become a problem once the constitution outgrows the maximum context length of the \textsc{Annotator} or \textsc{Decision} agent, which we do not evaluate in our work.




\begin{table}[t]
\centering
\small
\caption{Average prompt length across all datasets and models.}
\label{tab:method_averages}
\begin{tabular}{@{}lrrr@{}}
\toprule
\textbf{Metric} & \textbf{MAC} & \textbf{GEPA} & \textbf{MIPRO} \\
\midrule
Average Characters & 3,679.2 & 5,264.7 & 1,425.8 \\
\bottomrule
\end{tabular}
\end{table}

\section{Conclusion}
Constitutional AI offers transparent, inference-time control over LLM behavior, but writing effective constitutions is difficult and time-consuming. We formalize constitution learning as the task of automatically deriving interpretable rules from labeled examples, and propose MAC, a multi-agent method where specialized agents propose, critique, and refine rules.
MAC outperforms prompt-optimization baselines by over 50\% on private information tagging across legal, healthcare, and finance domains, and matches or exceeds pretrained taggers without task-specific training. Retrieval augmentation (reMAC) improves performance for specialized domains, while agent specialization (MAC+) enables smaller models to match larger ones without memorization risks. MAC also generalizes to tool calling, demonstrating applicability beyond classification.
We hope this work initiates research on constitution learning as a transparent and effective alternative to unstructured prompt optimization.

\section{Impact Statement}
All datasets used in our experiments are publicly available and contain either personally identifiable information (PII) that is authorized for public release or data shared with full consent from the relevant stakeholders. Any biases regarding what is considered private arise from the provided annotations, which reflect domain- or stakeholder-specific definitions. In addition, the underlying language models may encode biases from their pretraining data, which are outside the scope of this work and not modified by our approach.

This work lowers the barrier to deploying privacy-preserving language model systems. By enabling privacy behavior to be specified and adapted through explicit constitutions rather than model fine-tuning, stakeholders can mark and revise what is considered private using only rule updates, without retraining models or exposing sensitive data during optimization. This is particularly important in privacy-sensitive deployments, where fine-tuning can introduce memorization risks and operational constraints.

More broadly, our work advances a new paradigm of constitution learning, in which transparent, auditable rule sets are learned directly from data under task feedback. By restricting optimization to explicit rules rather than unconstrained prompt rewrites, this approach improves trust, interpretability, and stakeholder control while achieving strong performance. We believe this represents an important step toward safer and more accountable use of language models in high-stakes applications.

\section{Author Acknowledgment}
Francesco Pinto (affiliated with Google DeepMind) contributed in an individual advisory capacity.

\bibliography{sample}

@inproceedings{sang2003introduction,
  title={Introduction to the CoNLL-2003 shared task: Language-independent named entity recognition},
  author={Sang, Erik Tjong Kim and De Meulder, Fien},
  booktitle={Proceedings of the seventh conference on Natural language learning at HLT-NAACL 2003},
  pages={142--147},
  year={2003}
}

@article{brown2020language,
  title={Language models are few-shot learners},
  author={Brown, Tom and Mann, Benjamin and Ryder, Nick and Subbiah, Melanie and Kaplan, Jared D and Dhariwal, Prafulla and Neelakantan, Arvind and Shyam, Pranav and Sastry, Girish and Askell, Amanda and others},
  journal={Advances in neural information processing systems},
  volume={33},
  pages={1877--1901},
  year={2020}
}

@article{wei2022chain,
  title={Chain-of-thought prompting elicits reasoning in large language models},
  author={Wei, Jason and Wang, Xuezhi and Schuurmans, Dale and Bosma, Maarten and Xia, Fei and Chi, Ed and Le, Quoc V and Zhou, Denny and others},
  journal={Advances in neural information processing systems},
  volume={35},
  pages={24824--24837},
  year={2022}
}

@article{ouyang2022training,
  title={Training language models to follow instructions with human feedback},
  author={Ouyang, Long and Wu, Jeffrey and Jiang, Xu and Almeida, Diogo and Wainwright, Carroll and Mishkin, Pamela and Zhang, Chong and Agarwal, Sandhini and Slama, Katarina and Ray, Alex and others},
  journal={Advances in neural information processing systems},
  volume={35},
  pages={27730--27744},
  year={2022}
}

@inproceedings{du2023improving,
  title={Improving factuality and reasoning in language models through multiagent debate},
  author={Du, Yilun and Li, Shuang and Torralba, Antonio and Tenenbaum, Joshua B and Mordatch, Igor},
  booktitle={Forty-first International Conference on Machine Learning},
  year={2023}
}

@article{chan2023chateval,
  title={Chateval: Towards better llm-based evaluators through multi-agent debate},
  author={Chan, Chi-Min and Chen, Weize and Su, Yusheng and Yu, Jianxuan and Xue, Wei and Zhang, Shanghang and Fu, Jie and Liu, Zhiyuan},
  journal={arXiv preprint arXiv:2308.07201},
  year={2023}
}

@inproceedings{chen2024reconcile,
  title={Reconcile: Round-table conference improves reasoning via consensus among diverse llms},
  author={Chen, Justin and Saha, Swarnadeep and Bansal, Mohit},
  booktitle={Proceedings of the 62nd Annual Meeting of the Association for Computational Linguistics (Volume 1: Long Papers)},
  pages={7066--7085},
  year={2024}
}

@inproceedings{wang2025unveiling,
  title={Unveiling privacy risks in llm agent memory},
  author={Wang, Bo and He, Weiyi and Zeng, Shenglai and Xiang, Zhen and Xing, Yue and Tang, Jiliang and He, Pengfei},
  booktitle={Proceedings of the 63rd Annual Meeting of the Association for Computational Linguistics (Volume 1: Long Papers)},
  pages={25241--25260},
  year={2025}
}

@article{yan2025protecting,
  title={On protecting the data privacy of Large Language Models (LLMs) and LLM agents: A literature review},
  author={Yan, Biwei and Li, Kun and Xu, Minghui and Dong, Yueyan and Zhang, Yue and Ren, Zhaochun and Cheng, Xiuzhen},
  journal={High-Confidence Computing},
  pages={100300},
  year={2025},
  publisher={Elsevier}
}

@article{alizadeh2025simple,
  title={Simple Prompt Injection Attacks Can Leak Personal Data Observed by LLM Agents During Task Execution},
  author={Alizadeh, Meysam and Samei, Zeynab and Stetsenko, Daria and Gilardi, Fabrizio},
  journal={arXiv preprint arXiv:2506.01055},
  year={2025}
}

@inproceedings{akbik2019flair,
  title={FLAIR: An easy-to-use framework for state-of-the-art NLP},
  author={Akbik, Alan and Bergmann, Tanja and Blythe, Duncan and Rasul, Kashif and Schweter, Stefan and Vollgraf, Roland},
  booktitle={Proceedings of the 2019 conference of the North American chapter of the association for computational linguistics (demonstrations)},
  pages={54--59},
  year={2019}
}

@inproceedings{schmitt2019replicable,
  title={A replicable comparison study of NER software: StanfordNLP, NLTK, OpenNLP, SpaCy, Gate},
  author={Schmitt, Xavier and Kubler, Sylvain and Robert, J{\'e}r{\'e}my and Papadakis, Mike and LeTraon, Yves},
  booktitle={2019 sixth international conference on social networks analysis, management and security (SNAMS)},
  pages={338--343},
  year={2019},
  organization={IEEE}
}

@inproceedings{kaur2024bert,
  title={Bert-ner: A transformer-based approach for named entity recognition},
  author={Kaur, Navjeet and Saha, Ashish and Swami, Makul and Singh, Muskan and Dalal, Ravi},
  booktitle={2024 15th international conference on computing communication and networking technologies (ICCCNT)},
  pages={1--7},
  year={2024},
  organization={IEEE}
}

@article{li2020survey,
  title={A survey on deep learning for named entity recognition},
  author={Li, Jing and Sun, Aixin and Han, Jianglei and Li, Chenliang},
  journal={IEEE transactions on knowledge and data engineering},
  volume={34},
  number={1},
  pages={50--70},
  year={2020},
  publisher={IEEE}
}

@inproceedings{huang2024collective,
  title={Collective constitutional ai: Aligning a language model with public input},
  author={Huang, Saffron and Siddarth, Divya and Lovitt, Liane and Liao, Thomas I and Durmus, Esin and Tamkin, Alex and Ganguli, Deep},
  booktitle={Proceedings of the 2024 ACM Conference on Fairness, Accountability, and Transparency},
  pages={1395--1417},
  year={2024}
}

@misc{huang2022largepretrainedlanguagemodels,
      title={Are Large Pre-Trained Language Models Leaking Your Personal Information?}, 
      author={Jie Huang and Hanyin Shao and Kevin Chen-Chuan Chang},
      year={2022},
      eprint={2205.12628},
      archivePrefix={arXiv},
      primaryClass={cs.CL},
      url={https://arxiv.org/abs/2205.12628}, 
}

@misc{kim2023propileprobingprivacyleakage,
      title={ProPILE: Probing Privacy Leakage in Large Language Models}, 
      author={Siwon Kim and Sangdoo Yun and Hwaran Lee and Martin Gubri and Sungroh Yoon and Seong Joon Oh},
      year={2023},
      eprint={2307.01881},
      archivePrefix={arXiv},
      primaryClass={cs.CR},
      url={https://arxiv.org/abs/2307.01881}, 
}

@inproceedings{patil2025bfcl,
title={The Berkeley Function Calling Leaderboard (BFCL): From Tool Use to Agentic Evaluation of Large Language Models}, 
author={Patil, Shishir G. and Mao, Huanzhi and Cheng-Jie Ji, Charlie and Yan, Fanjia and Suresh, Vishnu and Stoica, Ion and E. Gonzalez, Joseph},
booktitle={Forty-second International Conference on Machine Learning},
year={2025},
}

@article{PII_EXP_1,
  title={The PII problem: Privacy and a new concept of personally identifiable information},
  author={Schwartz, Paul M and Solove, Daniel J},
  journal={NYUL rev.},
  volume={86},
  pages={1814},
  year={2011},
  publisher={HeinOnline}
}

@book{PII_EXP_2,
  title={Guide to protecting the confidentiality of personally identifiable information},
  author={McCallister, Erika},
  year={2010},
  publisher={Diane Publishing}
}

@article{ramnath2025systematic,
  title={A systematic survey of automatic prompt optimization techniques},
  author={Ramnath, Kiran and Zhou, Kang and Guan, Sheng and Mishra, Soumya Smruti and Qi, Xuan and Shen, Zhengyuan and Wang, Shuai and Woo, Sangmin and Jeoung, Sullam and Wang, Yawei and others},
  journal={arXiv preprint arXiv:2502.16923},
  year={2025}
}

@misc{qwen2025qwen25technicalreport,
      title={Qwen2.5 Technical Report}, 
      author={Qwen and : and An Yang and Baosong Yang and Beichen Zhang and Binyuan Hui and Bo Zheng and Bowen Yu and Chengyuan Li and Dayiheng Liu and Fei Huang and Haoran Wei and Huan Lin and Jian Yang and Jianhong Tu and Jianwei Zhang and Jianxin Yang and Jiaxi Yang and Jingren Zhou and Junyang Lin and Kai Dang and Keming Lu and Keqin Bao and Kexin Yang and Le Yu and Mei Li and Mingfeng Xue and Pei Zhang and Qin Zhu and Rui Men and Runji Lin and Tianhao Li and Tianyi Tang and Tingyu Xia and Xingzhang Ren and Xuancheng Ren and Yang Fan and Yang Su and Yichang Zhang and Yu Wan and Yuqiong Liu and Zeyu Cui and Zhenru Zhang and Zihan Qiu},
      year={2025},
      eprint={2412.15115},
      archivePrefix={arXiv},
      primaryClass={cs.CL},
      url={https://arxiv.org/abs/2412.15115}, 
}

@misc{nvidia_glinerPII_2025,
  title        = {{GLiNER-PII: PII/PHI Detection Model}},
  author       = {{NVIDIA}},
  year         = {2025},
  howpublished = {\url{https://huggingface.co/nvidia/gliner-PII}},
  note         = {Accessed: 2026-01-05; Token classification model for detecting personally identifiable information (PII) and protected health information (PHI) built on the GLiNER architecture.},
}

@misc{gliner,
      title={GLiNER2: An Efficient Multi-Task Information Extraction System with Schema-Driven Interface}, 
      author={Urchade Zaratiana and Gil Pasternak and Oliver Boyd and George Hurn-Maloney and Ash Lewis},
      year={2025},
      eprint={2507.18546},
      archivePrefix={arXiv},
      primaryClass={cs.CL},
      url={https://arxiv.org/abs/2507.18546}, 
}

@misc{GEPA,
      title={GEPA: Reflective Prompt Evolution Can Outperform Reinforcement Learning}, 
      author={Lakshya A Agrawal and Shangyin Tan and Dilara Soylu and Noah Ziems and Rishi Khare and Krista Opsahl-Ong and Arnav Singhvi and Herumb Shandilya and Michael J Ryan and Meng Jiang and Christopher Potts and Koushik Sen and Alexandros G. Dimakis and Ion Stoica and Dan Klein and Matei Zaharia and Omar Khattab},
      year={2025},
      eprint={2507.19457},
      archivePrefix={arXiv},
      primaryClass={cs.CL},
      url={https://arxiv.org/abs/2507.19457}, 
}

@misc{MIPRO,
      title={Optimizing Instructions and Demonstrations for Multi-Stage Language Model Programs}, 
      author={Krista Opsahl-Ong and Michael J Ryan and Josh Purtell and David Broman and Christopher Potts and Matei Zaharia and Omar Khattab},
      year={2024},
      eprint={2406.11695},
      archivePrefix={arXiv},
      primaryClass={cs.CL},
      url={https://arxiv.org/abs/2406.11695}, 
}

@misc{SFT,
      title={LoRA: Low-Rank Adaptation of Large Language Models}, 
      author={Edward J. Hu and Yelong Shen and Phillip Wallis and Zeyuan Allen-Zhu and Yuanzhi Li and Shean Wang and Lu Wang and Weizhu Chen},
      year={2021},
      eprint={2106.09685},
      archivePrefix={arXiv},
      primaryClass={cs.CL},
      url={https://arxiv.org/abs/2106.09685}, 
}

@misc{GRPO,
      title={DeepSeekMath: Pushing the Limits of Mathematical Reasoning in Open Language Models}, 
      author={Zhihong Shao and Peiyi Wang and Qihao Zhu and Runxin Xu and Junxiao Song and Xiao Bi and Haowei Zhang and Mingchuan Zhang and Y. K. Li and Y. Wu and Daya Guo},
      year={2024},
      eprint={2402.03300},
      archivePrefix={arXiv},
      primaryClass={cs.CL},
      url={https://arxiv.org/abs/2402.03300}, 
}

@inproceedings{li2025papillon,
  title={Papillon: Privacy preservation from internet-based and local language model ensembles},
  author={Li, Siyan and Raghuram, Vethavikashini Chithrra and Khattab, Omar and Hirschberg, Julia and Yu, Zhou},
  booktitle={Proceedings of the 2025 Conference of the Nations of the Americas Chapter of the Association for Computational Linguistics: Human Language Technologies (Volume 1: Long Papers)},
  pages={3371--3390},
  year={2025}
}

@inproceedings{yang2023large,
  title={Large language models as optimizers},
  author={Yang, Chengrun and Wang, Xuezhi and Lu, Yifeng and Liu, Hanxiao and Le, Quoc V and Zhou, Denny and Chen, Xinyun},
  booktitle={The Twelfth International Conference on Learning Representations},
  year={2023}
}

@article{staab2024large,
  title={Large language models are advanced anonymizers},
  author={Staab, Robin and Vero, Mark and Balunovi{\'c}, Mislav and Vechev, Martin},
  journal={arXiv preprint arXiv:2402.13846},
  year={2024}
}

@article{LORA,
  title={Lora: Low-rank adaptation of large language models},
  author={Hu, Edward J and Shen, Yelong and Wallis, Phillip and Allen-Zhu, Zeyuan and Li, Yuanzhi and Wang, Shean and Wang, Lu and Chen, Weizhu},
  journal={arXiv preprint arXiv:2106.09685},
  year={2021}
}

@inproceedings{echr_dataset,
    title = "Neural Legal Judgment Prediction in {E}nglish",
    author = "Chalkidis, Ilias  and
      Androutsopoulos, Ion  and
      Aletras, Nikolaos",
    editor = "Korhonen, Anna  and
      Traum, David  and
      M{\`a}rquez, Llu{\'\i}s",
    booktitle = "Proceedings of the 57th Annual Meeting of the Association for Computational Linguistics",
    month = jul,
    year = "2019",
    address = "Florence, Italy",
    publisher = "Association for Computational Linguistics",
    url = "https://aclanthology.org/P19-1424",
    doi = "10.18653/v1/P19-1424",
    pages = "4317--4323",
}

@misc{maccrobat2020,
  author       = {Caufield, J. Harry},
  title        = {MACCROBAT2020 Dataset},
  howpublished = {\url{https://doi.org/10.6084/m9.figshare.9764942.v2}},
  year         = {2020},
  note         = {Version 2 of the MACCROBAT2018 dataset on Figshare},
}

@misc{microsoft_presidio_docs,
  title        = {Presidio: Data Protection and De-identification SDK},
  howpublished = {\url{https://microsoft.github.io/presidio/}},
  author       = {{Microsoft}},
  year         = {2025},
  note         = {Accessed: 2025-08-28},
}

@article{bai2022constitutional,
  title={Constitutional ai: Harmlessness from ai feedback},
  author={Bai, Yuntao and Kadavath, Saurav and Kundu, Sandipan and Askell, Amanda and Kernion, Jackson and Jones, Andy and Chen, Anna and Goldie, Anna and Mirhoseini, Azalia and McKinnon, Cameron and others},
  journal={arXiv preprint arXiv:2212.08073},
  year={2022}
}

@article{sharma2025constitutional,
  title={Constitutional classifiers: Defending against universal jailbreaks across thousands of hours of red teaming},
  author={Sharma, Mrinank and Tong, Meg and Mu, Jesse and Wei, Jerry and Kruthoff, Jorrit and Goodfriend, Scott and Ong, Euan and Peng, Alwin and Agarwal, Raj and Anil, Cem and others},
  journal={arXiv preprint arXiv:2501.18837},
  year={2025}
}

@misc{carlini2023quantifyingmemorizationneurallanguage,
      title={Quantifying Memorization Across Neural Language Models}, 
      author={Nicholas Carlini and Daphne Ippolito and Matthew Jagielski and Katherine Lee and Florian Tramer and Chiyuan Zhang},
      year={2023},
      eprint={2202.07646},
      archivePrefix={arXiv},
      primaryClass={cs.LG},
      url={https://arxiv.org/abs/2202.07646}, 
}
\bibliographystyle{plainnat}

\section{Appendix}

\subsection{Dataset Statistics}
\label{app:dataset_stats}
The full statistics for the Legal, Healthcare and finance datasets used as part of this work are available in Table \ref{tab:dataset_stats}.

\begin{table*}[t]
\centering
\small
\caption{Dataset statistics across domains after preprocessing and subsampling. Statistics are reported over the final 192 documents.}
\label{tab:dataset_stats}
\begin{tabular}{@{}lccc@{}}
\toprule
\textbf{Statistic} & \textbf{MACCROBAT (Healthcare)} & \textbf{PUPA (Finance)} & \textbf{ECHR (Legal)} \\
\midrule
Total characters & 542{,}124 & 400{,}041 & 1{,}547{,}159 \\
Total private characters & 297{,}616 & 53{,}701 & 158{,}007 \\
Private characters (\%) & 54.9 & 13.42 & 10.21 \\
Public characters & 244{,}508 & 346{,}340 & 1{,}389{,}152 \\
Public characters (\%) & 45.1 & 86.58 & 89.79 \\
\midrule
Total private entities & 24{,}155 & 4{,}564 & 11{,}156 \\
Total private entity groups & 41 & 29 & 8 \\
Average entities per privacy group & 589.15 & 157.38 & 1{,}394.50 \\
Average characters per privacy group & 7{,}258.93 & 1{,}851.76 & 19{,}750.88 \\
Average characters per entity & 12.32 & 11.77 & 14.16 \\
\bottomrule
\end{tabular}
\end{table*}

\subsection{Agent Prompts}
The full system prompts used for each agent are lengthy and are omitted for brevity. We instead summarize the key logic and responsibilities of each agent below.

\subsubsection{Annotator Agent}
\label{app:annotator}

The \texttt{Annotator} agent applies the current constitution to perform span-level private information tagging. Its system prompt defines the agent as a constitutional privacy annotation model and instructs it to apply the natural-language rules \emph{systematically} to identify all private phrases in a given document, targeting high precision and recall.

The user prompt provides (i) the full constitution text as a list of rules and (ii) the document to be annotated. The agent is required to return a strict JSON object of the form \texttt{\{"private\_phrases": [...]\}}, where each entry is a text span extracted from the document. The prompt emphasizes that, when uncertain, the agent should prefer marking a span as private, explicitly prioritizing low false negatives (FN) over low false positives (FP), consistent with privacy-preserving deployment.

To improve generalization, the prompt instructs the agent to use examples embedded in the rules as guidance but not to rely on examples as an exhaustive list. Instead, the agent must extrapolate from the rule descriptions to capture semantically similar spans. If the constitution is empty or unavailable, the agent is instructed to fall back to its own estimate of what constitutes private information, ensuring the system remains functional even when rule learning has not yet produced a stable constitution.

\subsubsection{Decision Agent}
\label{app:decision}
The \texttt{Decision} agent is the core control component that selects how to update the constitution based on observed tagging errors. Its system prompt specifies that it must (i) inspect the current constitution rules used by the annotator, (ii) analyze the current validation error profile, and (iii) choose a strategic action to update the rule set: \textsc{Add}, \textsc{Edit}, or \textsc{Remove}. The prompt explicitly frames the decision as optimizing downstream tagging robustness while keeping the constitution concise.

The user prompt provides a structured snapshot of the current state, including counts of false negatives (FN) and false positives (FP), total errors, the FP-per-FN ratio, the current number of rules, and optional trend information over recent updates. In addition, the agent is shown representative error patterns as sampled FN phrases (missed private spans) and FP phrases (incorrectly flagged spans), together with the full list of current constitutional rules. This structure forces the agent to ground decisions in concrete failure cases rather than abstract policy.

The prompt then guides the agent through three checks: (1) assess whether FN or FP is the dominant issue by examining both the counts and the semantic patterns in the sampled phrases; (2) use the trend signal to detect whether recent changes are making rules overly broad (e.g., FP increasing) or overly conservative (e.g., persistent FN); and (3) verify whether current rules already cover the observed FN patterns (suggesting \textsc{Edit}) or whether there are clear gaps (suggesting \textsc{Add}), as well as whether any rule appears to be responsible for systematic FP (suggesting \textsc{Remove} or \textsc{Edit}).

Action selection is constrained by explicit heuristics. The agent is instructed to choose \textsc{Remove} when FP dominates or when the constitution becomes overly verbose, and to output the specific rule index to remove. It is instructed to choose \textsc{Add} when FN is the primary issue and FP remains controlled, provided the system has capacity to introduce additional rules. It is instructed to choose \textsc{Edit} when existing rules appear misaligned with the observed errors, e.g., a rule is too broad and causes FP, or too narrow and causes FN, and to output the rule index to edit. The prompt further prioritizes recall by explicitly instructing the agent to prefer reducing FN over reducing FP, reflecting the task objective that missing private spans is worse than over-redacting.

Finally, the agent is required to return a strict JSON object containing: the chosen action, a rule index (or $-1$ for \textsc{Add}), and a short step-by-step rationale. Enforcing a fixed JSON schema ensures the decision is machine-readable and can be routed deterministically to the \texttt{Creator} or \texttt{Editor} agent for the subsequent update step.

\subsubsection{RuleCreator Agent}
\label{app:creator}
The \texttt{Creator} agent is responsible for proposing new constitutional rules when the decision agent selects an \textsc{Add} action. Its system prompt frames the agent as an expert constitutional rule engineer whose objective is to improve private information tagging performance by introducing a single, well-targeted rule grounded in observed failure patterns.

The prompt provides the agent with several structured inputs. First, it includes the prior decision analysis, which summarizes the reasoning that led to the choice of adding a rule, ensuring the new rule is aligned with the identified failure mode. Second, the current constitution state is provided, allowing the agent to avoid duplicating or overlapping existing rules. Third, the agent is shown sampled false negatives and false positives together with their surrounding context. These error examples are explicitly presented as partial evidence rather than exhaustive cases, discouraging memorization and encouraging generalization. Finally, a learning history of previously rejected updates is included so that the agent can avoid repeating ineffective rule formulations.

The prompt imposes strict constraints on rule construction. The agent must output exactly one new rule, expressed as concise natural language under a fixed length budget, and must not include literal private tokens that would overfit to the observed data. Rules are required to clearly specify what should be marked as private, followed by boundaries that clarify what should not be marked. To support flexibility, the prompt enumerates multiple permissible rule-writing styles (e.g., direct, contrast-based, context-triggered, function-based), encouraging the agent to select or combine styles that best address the observed errors rather than mimicking existing rules.

The agent is required to return its output as a strict JSON object containing only the proposed rule text. This constraint ensures the output can be programmatically inserted into the constitution and subsequently evaluated, while preserving interpretability and auditability of the learned rules.

\subsubsection{RuleEditor Agent}
\label{app:editor}
The \texttt{Editor} agent is invoked when the decision agent selects an \textsc{Edit} action. Its role is to revise a specific existing rule in the constitution to correct observed failure modes while preserving the overall structure and intent of the rule set. Unlike the creator agent, which introduces new rules, the editor agent performs targeted modifications to an identified rule.

The prompt provides the agent with the index of the rule to be edited and the full current constitution, allowing the agent to localize the change and avoid introducing redundancy. The agent is also given the prior decision reasoning explaining why this rule was selected for editing, grounding the update in the strategic diagnosis of the system’s errors. In addition, the agent receives sampled false negatives and false positives together with their surrounding context, which it must analyze to understand how the existing rule contributes to over- or under-marking.

To prevent ineffective updates, the prompt includes a history of previously rejected edits based on validation F1, encouraging the agent to avoid repeating unsuccessful modifications. The agent is instructed to make \emph{surgical} edits, refining boundaries, conditions, or scope rather than rewriting the rule wholesale, unless a change in style is warranted. Multiple rule-writing styles are permitted, but all edits must remain concise, avoid memorization of literal private tokens, and stay within a fixed length budget.

The editor agent is required to output exactly one revised rule as a strict JSON object. This constraint ensures that edits are directly comparable, auditable, and can be automatically substituted into the constitution for subsequent evaluation.

\subsection{Implementation}
\label{app:baseline_details}

For all evaluations, we use Qwen~2.5 Instruct models \cite{qwen2025qwen25technicalreport} at 3B, 7B, and 14B parameters, deployed with vLLM. Inference uses temperature 1.0, sampling enabled, $\texttt{top\_p}=0.9$, $\texttt{top\_k}=50$, and repetition penalty 1.0. We report F1 as the primary metric, computed by exact matching between predicted and ground-truth private spans, where only exact span matches are counted as correct. We compare MAC against three baselines: pretrained PII taggers used in practice, prompt-optimization methods that optimize natural-language instructions without updating model parameters, and training-based methods including supervised fine-tuning \cite{SFT} and GRPO \cite{GRPO}. For all approaches, training is run for 3 epochs over 20 batches, with validation performed every 5 batches.

\textbf{Pretrained taggers:}
We compare against commonly used non-generative PII tagging models. As a representative baseline, we use Microsoft Presidio \cite{microsoft_presidio_docs}, a modular framework combining rule-based methods, NER \cite{li2020survey}, and context-aware detectors, and adopted in prior work \cite{staab2024large}. We evaluate its best-performing components used in practice: BERT-NER \cite{kaur2024bert}, SpaCy \cite{schmitt2019replicable}, and Flair \cite{akbik2019flair}, which perform span-level tagging with fixed entity definitions. We also benchmark GLiNER-PII \cite{nvidia_glinerPII_2025}, a non-generative span-level model that supports over 55 PII and PHI categories across structured and unstructured text.

\textbf{Prompt optimization baselines:}
We compare against Genetic-Pareto (GEPA) \cite{GEPA}, which performs evolutionary prompt search using task feedback, and Multiprompt Instruction Proposal Optimizer (MIPRO) \cite{MIPRO}, which iteratively proposes and selects prompt variants. MAC uses 20 batches of 8 documents over 3 epochs, totaling 480 forward passes, and GEPA and MIPRO are configured with the same budget for fairness. For GEPA, we set $ \texttt{max\_metric\_calls}=480$, $\texttt{batch\_size}=8$, and use a reflection model with $\texttt{max\_context\_length}=3000$ and $\texttt{temperature}=1.0$. For MIPRO, we cap optimization at 480 total evaluations for parity. We use $\texttt{batch\_size}=8$, run 60 trials with full evaluation every 5 steps, and consider 32 candidate prompts per iteration. The reflection and evaluation models use a $\texttt{max\_context\_length}=3000$ and $\texttt{temperature}=1.0$, with minibatch-based evaluations over 8 documents.

\textbf{Training-Based Baselines:}

For GRPO, we train with 8 rollouts per prompt, a maximum completion length of 512 tokens, and a maximum prompt length of 3000 tokens. Training uses a per-device batch size of 8 for 3 epochs, with learning rate $1\times10^{-5}$, a cosine scheduler, and warmup ratio 0.1. Sampling uses temperature 1.0 and top-$p$ 0.9, with no KL regularization ($\beta=0.0$). To match data exposure, we sample 20 training examples per epoch and generate 8 rollouts per example, yielding 160 rollouts per epoch. GRPO uses the same LoRA configuration as SFT.

For SFT, we fine-tune using Low-Rank Adaptation (LoRA) \cite{LORA} with rank $r=64$, scaling factor $\alpha=128$, and dropout 0.05, applied to attention and feed-forward projection layers (\texttt{q\_proj}, \texttt{k\_proj}, \texttt{v\_proj}, \texttt{o\_proj}, \texttt{up\_proj}, \texttt{down\_proj}, \texttt{gate\_proj}). Training uses the same 20 batches for 3 epochs, with validation and testing on the same 16-document splits used across all methods.

\subsection{Constitutions learnt using MAC for all model sizes and datasets}
\label{app:mac_constitutions}
The first 3 rules from MAC learnt constitutions are available in Table \ref{tab:learned_constitutions_fullpage}.
\begin{table*}[t]
\centering
\small
\setlength{\tabcolsep}{6pt}
\renewcommand{\arraystretch}{1.25}
\newcolumntype{Y}{>{\raggedright\arraybackslash}X}
\caption{Learned constitutions (top 3 rules) for each model size and dataset. Each cell lists the first three rules of the best-selected constitution.}
\label{tab:learned_constitutions_fullpage}
\begin{tabularx}{\textwidth}{@{}l Y Y Y@{}}
\toprule
\textbf{Model} & \textbf{Legal} & \textbf{Healthcare} & \textbf{Finance} \\
\midrule



\textit{3B} & \textit{\# Rules: 3} & \textit{\# Rules: 7} & \textit{\# Rules: 13} \\
\specialrule{0.08em}{0.2em}{0.4em}

\textbf{Rules} &
\textbf{1.} Mark names, dates, OTPs, IDs, phone numbers; exclude non-private terms (e.g., ``applicant'', months); mark names only when supporting a specific private item/ID; exclude non-identifiers (e.g., ``in'', ``and'', ``on'').\par
\textbf{2.} Mark numeric dates; do not mark surrounding words. Example: 1982, 15 December 2004.\par
\textbf{3.} Mark dates (YYYY, MM/DD/YYYY); exclude common non-private terms (e.g., ``the applicant'', ``court''); do not mark ordinal suffixes (``st'', ``nd'', ``rd'', ``th''). &

\textbf{1.} Mark personal identifiers (names, dates, SSNs, phone numbers, emails, etc.); do not mark common words/placeholders (e.g., ``IV'', ``hands'', ``died'').\par
\textbf{2.} Mark full names, dates, SSNs, phone numbers, emails, medical IDs (ICD/MRN), lab test IDs (PT\#, CBC\#), and identifiers in context (e.g., ``ID:'', ``DOB:''); do not mark clinical descriptors unless part of the identifier.\par
\textbf{3.} Mark full names, DOBs, SSNs, phone numbers, emails, medical/lab IDs and labeled identifiers; do not mark generic clinical terms (e.g., ``progression'', ``recovery'') unless tied to specific identifiers. &

\textbf{1.} Mark tokens that uniquely identify individuals (e.g., names, emails, phone numbers, SSNs, addresses, dates); avoid generic context or logs unless they explicitly identify a person; exclude non-identifying descriptors and industry terms. Example: John Doe <j.doe@company.com> 1952-06-24 95 AVA BLANCET FRAUZ.\par
\textbf{2.} Mark explicit identifiers and labeled fields (e.g., Name, Email, Address, Date) when they contain identifying information; exclude generic narrative text and industry-specific terminology without identifiers. Example: John Doe <j.doe@company.com> 1952-06-24 95 AVA BLANCET FRAUZ.\par
\textbf{3.} Mark identifiers appearing in structured or semi-structured contexts (e.g., IDs, contact fields, combined identifiers); do not mark log-only or industry terms unless they directly reference an identifiable entity. \\

\addlinespace[0.6em]

\midrule
\textit{7B} & \textit{\# Rules: 4} & \textit{\# Rules: 12} & \textit{\# Rules: 16} \\
\specialrule{0.08em}{0.2em}{0.4em}

\textbf{Rules} &
\textbf{1.} Mark dates, times, and durations that specifically reference case numbers or dates of legal events; do not mark general temporal references or unrelated numbers.\par
\textbf{2.} Mark personal names; do not mark titles/roles unless they precede the name.\par
\textbf{3.} Mark nationality, titles following names, and specific government officials; do not mark generic titles or roles. &

\textbf{1.} Mark phrases indicating medical conditions or treatments (drug names, disease names, procedures), but not surrounding text unless it directly identifies a person.\par
\textbf{2.} Mark phrases indicating medical conditions/treatments and clinical descriptors (e.g., ``best-corrected visual acuity''); do not mark surrounding text unless directly identifying.\par
\textbf{3.} Mark age/gender/introduction phrases in records (e.g., ``62-year-old'', ``female'', ``presented to...''); do not mark surrounding text unless directly identifying. &

\textbf{1.} Mark dates, times, monetary values, and names as private; do not mark surrounding text unless it directly identifies a person.\par
\textbf{2.} Mark names as private; do not mark surrounding text unless it directly identifies a person.\par
\textbf{3.} Mark names, email addresses, and phone numbers as private; do not mark surrounding text unless it directly identifies a person. \\
\addlinespace[0.6em]

\midrule
\textit{14B} & \textit{\# Rules: 15} & \textit{\# Rules: 4} & \textit{\# Rules: 7} \\
\specialrule{0.08em}{0.2em}{0.4em}

\textbf{Rules} &
\textbf{1.} Mark as private specific dates and location names within legal/case documents; do not mark generic descriptors or full sentences containing such terms.\par
\textbf{2.} Mark official titles when they refer to individuals, specific dates, and specific location names; do not mark generic titles or broad locations.\par
\textbf{3.} Mark specific numbers in case details (e.g., witnesses, event dates) but not general narrative numbers. &

\textbf{1.} Mark medical procedures, diagnoses, and medical equipment in clinical contexts as private; do not mark general terms outside medical narratives; split adjacent phrases when needed.\par
\textbf{2.} Mark conditions, treatments, procedures, and anatomical features in medical contexts; do not mark general terms outside medical narratives; split adjacent phrases when needed.\par
\textbf{3.} Mark age, gender, and personal titles when appearing in referral/record introductions (e.g., ``A 68-year-old man...''). &

\textbf{1.} Context-triggered: mark dates/times/names/contact details when preceded by labels (e.g., ``Date:'', ``Name:''); do not mark the labels.\par
\textbf{2.} Mark any phrase that can uniquely identify an individual/organization (names, dates, contact details, identifiers); do not mark generic roles/broad locations/placeholders without identifiers.\par
\textbf{3.} Function-based: mark phrases containing multiple identifying elements in a single statement; avoid marking generic text unless it includes specific identifiers. \\
\bottomrule
\end{tabularx}
\end{table*}

\subsection{Re-MAC: Retrieval augmented MAC}
\label{app:remac_full_plots}
\begin{figure*}[t]
    \centering
    \includegraphics[width=\textwidth]{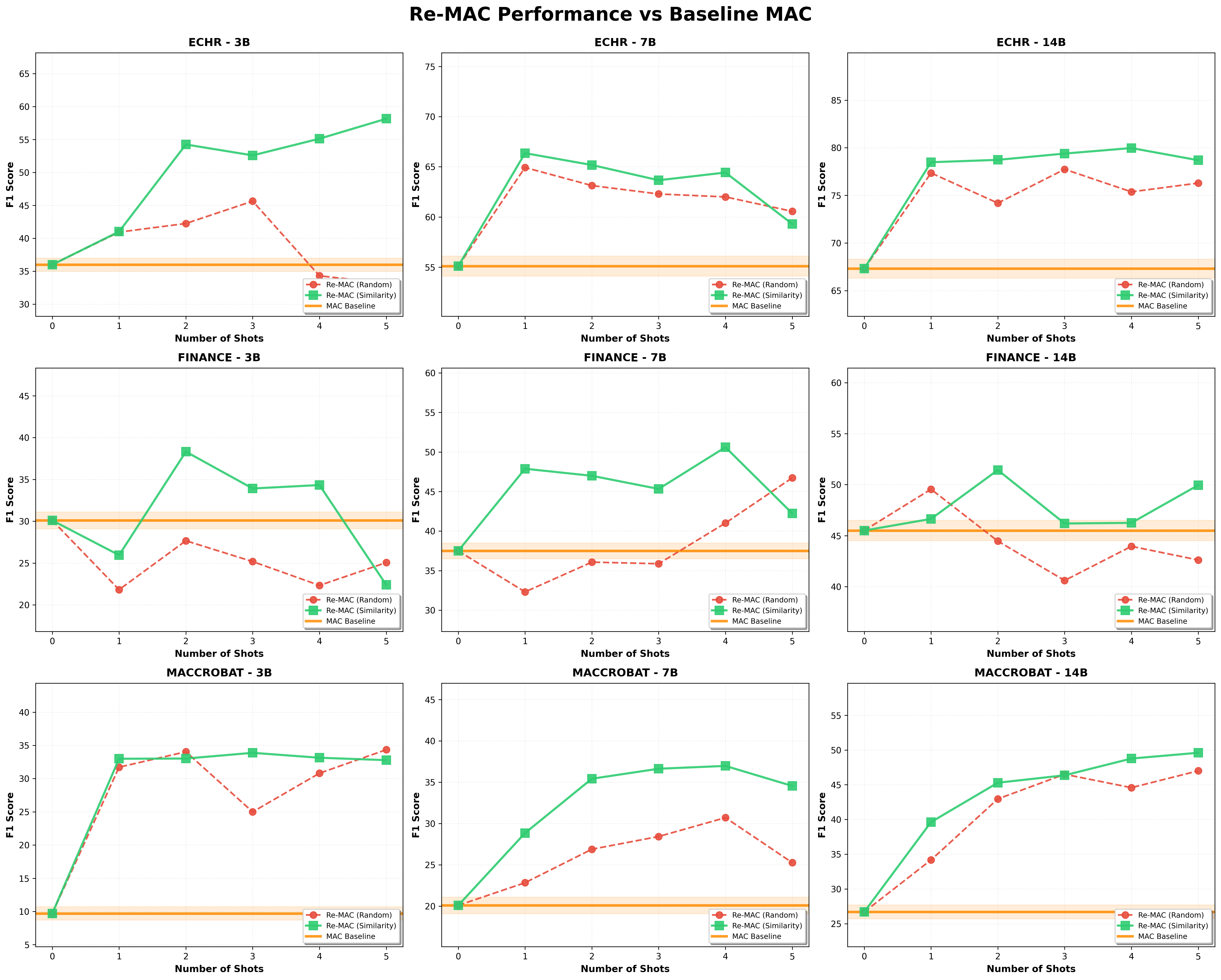}
    \caption{Re-MAC: Retrieval augmented MAC, test-performance (F1)
for different numbers of retrieved examples.}
    \label{fig:re-mac-plot}
\end{figure*}

Plots showcasing the improvement in performance by adding different few shot samples (documents) using different strategies in Re-MAC is shown in Figure \ref{fig:re-mac-plot}.
\subsection{Learning a constitution using MAC}
\label{app:constitutional_evolution}
\begin{figure*}[t]
    \centering
    \includegraphics[width=\textwidth]{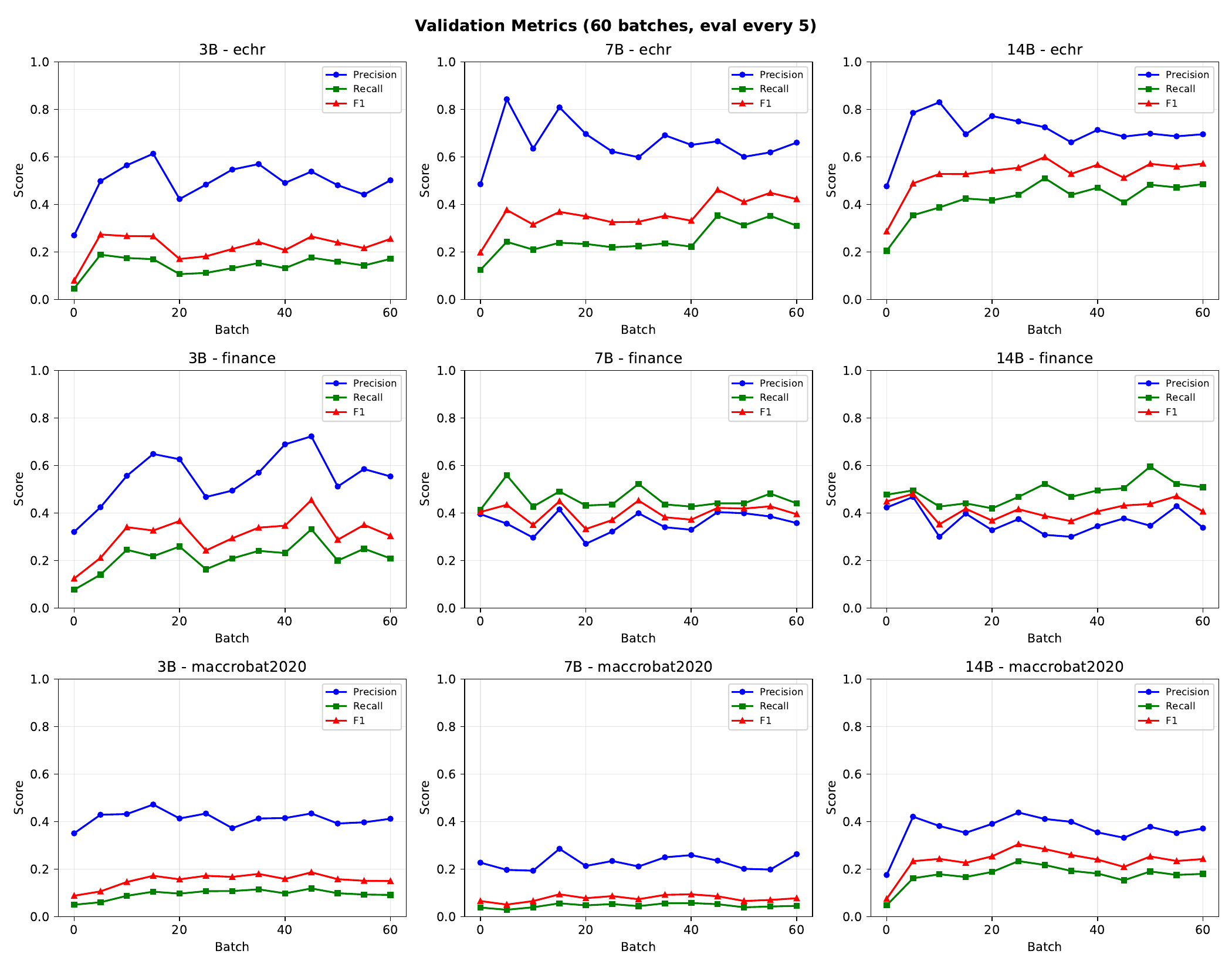}
    \caption{Validation precision, recall and F1 over batches for MAC across different model sizes.}
    \label{fig:val_f1_vs_batch}
\end{figure*}

Plots showcasing the evolution of validation F1 by updating the constitution are shown in Figure \ref{fig:val_f1_vs_batch}.

\subsection{MAC+ Loss Curves}
\label{app:mac_plus_loss_curves}
\begin{figure}
    \centering
    \includegraphics[width=\columnwidth]{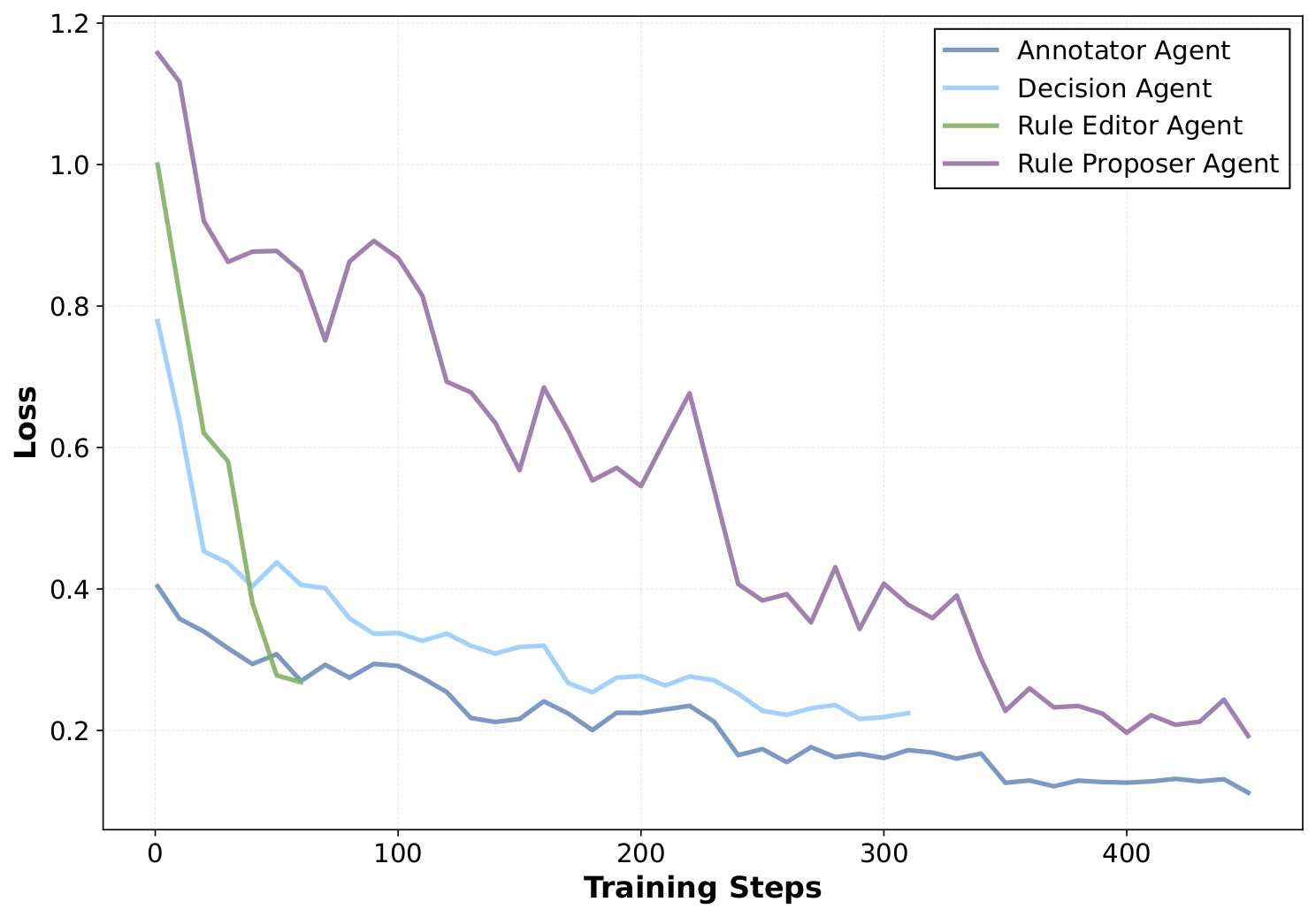}    
    \caption{Loss curves over training steps for all agents in MAC.}
    \label{fig:mac_plus_loss_curves}
\end{figure}

Loss curves for all agents tuned as part of MAC+ are shown in Figure~\ref{fig:mac_plus_loss_curves}.

\subsection{Rollout Statistics for Dataset used for MAC+}
\label{app:rollout_statistics}
Statistics for the dataset used to tune each agent in MAC to obtain MAC+ are shown in Table~\ref{tab:mac_plus_token_stats}. We collect agent-specific execution trajectories generated by the 14B model and use them to supervise and specialize each agent via fine-tuning. Less frequently invoked agents contribute all available trajectories, while more frequent agents are capped at 1,000 samples. 

\begin{table*}[tb]
\centering
\small
\setlength{\tabcolsep}{6pt}
\begin{tabular}{l r r r r r}
\toprule
Agent Role & Rollouts & Total Input & Total Output & Avg. Input & Avg. Output \\
\midrule
Annotator Agent        & 1{,}000 & 1{,}466{,}839 & 378{,}013 & 1{,}466.8 & 378.0 \\
Decision Agent         &   697   & 1{,}402{,}748 &  64{,}257 & 2{,}012.6 &  92.2 \\
Rule Editor Agent      &   132   &   366{,}927   &  11{,}327 & 2{,}779.8 &  85.8 \\
Rule Proposer Agent    & 1{,}000 & 3{,}178{,}303 &  74{,}905 & 3{,}178.3 &  74.9 \\
\midrule
Total                  & 2{,}829 & 6{,}414{,}817 & 528{,}502 & 2{,}267.5 & 186.8 \\
\bottomrule
\end{tabular}
\caption{Rollout statistics and token counts per agent role used for supervised fine-tuning in MAC+.}
\label{tab:mac_plus_token_stats}
\end{table*}

\subsection{Outputs from MAC Agents for a random rollout - Legal}
\label{app:sample_outputs_mac}
Sample outputs from the MAC Annotator Agent are shown in Table~\ref{tab:annotar_output}. Examples of Decision Agent actions are provided for adding rules (Table~\ref{app:decision_add}), removing rules (Table~\ref{app:decision_remove}), and editing rules (Table~\ref{app:decision_edit}).

\begin{table*}[t]
\centering
\caption{}
\begin{tcolorbox}[
  width=\textwidth,
  colback=blue!8,
  colframe=black,
  title=MAC Annotator agent output for a random rollout -- Legal,
  colbacktitle=blue!90!black,
  coltitle=white,
  fonttitle=\bfseries,
  enhanced,
  sharp corners=south
]
\label{tab:annotar_output}
\begin{verbatim}
Based on the provided constitution rules, here are the private phrases 
identified in the text:

```json
{
 "private_phrases": [
    "application (no. 48545/99)",
-------------------
REMOVED FOR BREVITY
-------------------
    "Court of Cassation",
    "15 November 1999"
 ]
}
```

### Explanation:
1. **Application Number**: `"application (no. 48545/99)"` - 
Specific application number, marked as private per Rule 21.
2. **Individual Names**: `"Mr Mehmet Şah Çelik"`, 
`"Mr Ferhan Arasan"`, `"Mr Mehmet Emin Güngör"` - Specific 
names of individuals in professional/legal contexts, marked
as private per Rules 6, 18, and 25.
3. **Dates**: 
   - `"16 December 1994"`, `"30 December 1994"`, 
   `"11 January 1995"`, `"17 January 1995"`, 
   `"16 March 1995"`, `"7 September 1995"`, 
   `"7 November 1996"`, `"19 December 1996"`, `"8 July 1997"`,
   `"18 December 1998"`, `"15 November 1999"` 
   - Specific dates in legal/professional contexts, 
   marked as private per Rules 5, 10, 17, and 27.
4. **Organizations**: `"Batman Security Directorate"`, 
`"Batman Magistrates’ Court"`, 
`"Diyarbakır State Security Court"`, 
`"Court of Cassation"` - Specific names of organizations 
directly linked to legal actions, marked as private per 
Rules 12, 16, and 25.
5. **Birth Year**: `"1979"` - Specific year of birth, 
marked as private per Rule 11.
6. **Prison Name**: `"Batman E-type Prison"`
- Specific name of prison, marked as private per Rule 13.
7. **Organization Name**: `"PKK (the Kurdish Workers’ Party)"
` - Specific name of an organization in a professional/legal 
context, marked as private per Rule 25.

These phrases are identified as private according 
to the provided rules.
\end{verbatim}
\end{tcolorbox}
\end{table*}

\begin{table*}[t]
\centering
\caption{}
\definecolor{decisionblue}{HTML}{99CCFF}
\definecolor{decisionblueDark}{HTML}{336699}

\begin{tcolorbox}[
  width=\textwidth,
  colback=decisionblue!20,
  colframe=black,
  title=MAC Decision agent output for a random rollout -- Legal (Add operation),
  colbacktitle=decisionblueDark,
  coltitle=white,
  fonttitle=\bfseries,
  enhanced,
  sharp corners=south
]

\begin{verbatim}
{
  "action": "add",
  "rule_index": "-1",
  "reasoning": "Step-by-step analysis: [1] Current balance assessment: There
  are significantly more False Negatives compared to False Positives. [2]
  Main problem identified: The system is missing important privacy terms 
  (high FN). [3] Why this action best addresses it: Adding new rules to 
  cover missed privacy terms will improve the system's accuracy without 
  causing an excessive number of false alarms."
}
\end{verbatim}
\label{app:decision_add}
\end{tcolorbox}
\end{table*}

\begin{table*}[t]
\centering
\caption{}
\definecolor{decisionblue}{HTML}{99CCFF}
\definecolor{decisionblueDark}{HTML}{336699}

\begin{tcolorbox}[
  width=\textwidth,
  colback=decisionblue!20,
  colframe=black,
  title=MAC Decision agent output for a random rollout -- Legal (Remove operation),
  colbacktitle=decisionblueDark,
  coltitle=white,
  fonttitle=\bfseries,
  enhanced,
  sharp corners=south
]

\begin{verbatim}
{
  "action": "remove",
  "rule_index": 0,
  "reasoning": "Current balance assessment: The system is marking too much 
  as private, as evidenced by the high number of false positives (FP) 
  compared to false negatives (FN). Main problem identified: The rules are
  overly broad, matching non-private terms like dates and places in the FP
  list. Why this action best addresses it: Removing the first rule will 
  reduce the likelihood of false alarms while allowing for more targeted 
  rules to address the FN patterns."
}
\end{verbatim}
\label{app:decision_remove}
\end{tcolorbox}
\end{table*}

\begin{table*}[t]
\centering
\definecolor{decisionblue}{HTML}{99CCFF}
\definecolor{decisionblueDark}{HTML}{336699}
\caption{}
\begin{tcolorbox}[
  width=\textwidth,
  colback=decisionblue!20,
  colframe=black,
  title=MAC Decision agent output for a random rollout -- Legal (Edit operation),
  colbacktitle=decisionblueDark,
  coltitle=white,
  fonttitle=\bfseries,
  enhanced,
  sharp corners=south
]

\begin{verbatim}
{
  "action": "edit",
  "rule_index": 0,
  "reasoning": "Step-by-step analysis: [1] Current balance
  assessment: FN phrases are general temporal references 
  like '1 May 1993', '18 years', and '6 September 2007', 
  while FP phrases include specific legal event references
  such as '26 April 1993'. This suggests the system is too
  conservative. [2] Main problem identified: The system is
  missing privacy items due to overly specific rules, 
  leading to a high number of FN errors. [3] Why this action
  best addresses it: Editing Rule 0 to include general 
  temporal references will help reduce FN errors while 
  maintaining control over FP."
}
\end{verbatim}
\label{app:decision_edit}
\end{tcolorbox}
\end{table*}

\subsection{Learnt prompts for GEPA - Legal}
\label{app:gepa_prompts}

GEPA-learned prompts for the Annotator Agent on the ECHR dataset are provided as follows: 3B model prompts in Tables~\ref{app:gepa_prompt_3b} and \ref{app:gepa_prompt_3b_contd}; 7B model prompts in Tables~\ref{app:gepa_prompt_7b}, \ref{app:gepa_prompt_7b_contd}, and \ref{app:gepa_prompt_7b_contd_2}; and 14B model prompts in Tables~\ref{app:gepa_prompt_14b} and \ref{app:gepa_prompt_14b_contd}.

\begin{table*}[t]
\centering
\caption{}
\begin{tcolorbox}[
  width=\textwidth,
  colback=white!95!gray,
  colframe=gray!80!black,
  title=GEPA learnt prompt using a 3B model for private-token tagging on the Legal dataset,
  coltitle=white,
  fonttitle=\bfseries,
  colbacktitle=gray!80!black,
  enhanced,
  sharp corners=south
]
\begin{verbatim}
json
{
 "private_phrases": []
}
```
### Detailed Task Description

You are tasked with identifying and annotating private phrases within a 
given text related to the protection of human rights and fundamental 
freedoms as enshrined in the Constitution of Human Rights and Fundamental 
Freedoms (Convention for the Protection of Human Rights and
Fundamental Freedoms). The annotated phrases should reflect sensitive 
personal information, such as names, dates of birth, addresses, financial
details, and other potentially identifiable data that, if disclosed, could 
infringe upon an individual's rights 
and fundamental freedoms.

### Input Format
The input text will be a block of natural language describing factual
events and contexts related to court cases, personal information, and 
financial matters. Your task is to parse through this text and highlight
any phrases that could potentially identify individuals, groups, locations,
or other private entities.

### Annotations Format
Output should be in a JSON string as specified. The `private_phrases` 
field will contain an array of strings representing the phrases identified 
as private according to the guidelines above.

### Example Inputs
#### Input Text
```
"The case originated in an application (no. 3042/05) against the Republic
of Turkey lodged with the Court under Article 34 of the Convention for the
Protection of Human Rights and Fundamental Freedoms ("the Convention") by 
a Turkish national, Mr Nihat Konak ("the applicant"), on 19 November 2004. 
The applicant was represented by Mr M. A. Kırdök, a lawyer practising in 
Istanbul. The Turkish Government ("the Government") were represented by 
their Agent."
```
#### Expected Output
```json
{
 "private_phrases": [
  "Mr Nihat Konak",
  "Mr M. A. Kırdök",
  "Istanbul"
 ]
}
```
\end{verbatim}
\label{app:gepa_prompt_3b}
\end{tcolorbox}
\end{table*}

\begin{table*}[t]
\centering
\caption{}
\begin{tcolorbox}[
  width=\textwidth,
  colback=white!95!gray,
  colframe=gray!80!black,
  title=GEPA learnt prompt using a 3B model for private-token tagging on the Legal dataset (Continued...),
  coltitle=white,
  fonttitle=\bfseries,
  colbacktitle=gray!80!black,
  enhanced,
  sharp corners=south
]
\begin{verbatim}

### Constraints and Considerations
- **Accuracy**: Ensure that your annotations match the input text 
accurately and comply with the Constitution's guidelines for privacy.
- **Generalizability**: Focus on private entities rather than relying
on examples provided in previous feedback.

### Feedback Analysis
Feedback will consider the precision (how many false positives and
false negatives are there) and recall (how comprehensive are the 
annotations).

Your goal is to annotate the input text effectively and iteratively
improve your annotations based on feedback.

Adhere to the provided guidelines and JSON structure for the final
output.
\end{verbatim}
\label{app:gepa_prompt_3b_contd}
\end{tcolorbox}
\end{table*}

\begin{table*}[t]
\centering
\caption{}
\begin{tcolorbox}[
  width=\textwidth,
  colback=white!95!gray,
  colframe=gray!80!black,
  title=GEPA learnt prompt using a 7B model for private-token tagging on the Legal dataset,
  coltitle=white,
  fonttitle=\bfseries,
  colbacktitle=gray!80!black,
  enhanced,
  sharp corners=south
]
\begin{verbatim}

## Instructions:

You are a constitutional privacy annotation agent. Your task 
is to systematically apply a given set of constitutional rules
to identify and annotate private phrases within a provided text. 
The goal is to ensure maximum precision and recall by marking all
potentially private information, even if it means marking some 
non-private information as private to avoid missing any actual
private details. Follow the detailed task description and 
guidelines below to complete the task accurately.

### Detailed Task Description:

1. **Constitutional Rules:** You will be given rules in natural 
language that define what constitutes private information. Use 
these rules to guide your annotations. The rules may include 
examples, which you should understand and extrapolate upon to 
identify all matching private phrases.
2. **Annotation Task:** Apply the constitutional rules to extract
all matching private phrases from the provided text. The output 
should be structured as JSON, containing a list of all identified
private phrases.
3. **Precision and Recall:** Prioritize high recall over high
precision. This means that it is safer to err on the side of marking
something as private when unsure to avoid missing any potentially 
private information.
4. **Generalization:** If no specific constitutional rules are 
provided, you should use your judgment to estimate what would be 
considered private based on typical legal and ethical standards. 
Do not simply rely on memorized examples.
5. **Feedback Loop:** After completing the task, a feedback score will
be provided (F1 score). Use this feedback to refine your approach and
improve the accuracy of your annotations.

### Guidelines:

- **Understand the Constitutional Rules Thoroughly:** Carefully
read and comprehend the given constitutional rules. Extrapolate 
from examples to cover all potential private information.
- **Extrapolate from Examples:** Use provided examples to infer 
patterns and apply them to unseen texts.
- **Prioritize Marking as Private:** If unsure, mark something as 
private to avoid missing any potentially private information.
- **Use General Legal and Ethical Standards:** If no constitutional
rules are provided, apply general legal and ethical standards to 
estimate what would be considered private.
- **Continuous Improvement:** Use the feedback (F1 score) to adjust
and refine your approach for better precision and recall.

\end{verbatim}
\label{app:gepa_prompt_7b}
\end{tcolorbox}
\end{table*}

\begin{table*}[t]
\centering
\caption{}
\begin{tcolorbox}[
  width=\textwidth,
  colback=white!95!gray,
  colframe=gray!80!black,
  title=GEPA learnt prompt using a 7B model for private-token tagging on the Legal dataset (Continued...),
  coltitle=white,
  fonttitle=\bfseries,
  colbacktitle=gray!80!black,
  enhanced,
  sharp corners=south
]
\begin{verbatim}
### Input Format:
- **clinical_text:** The text to be annotated for privacy.

### Output Format:
- JSON with a list of all identified private phrases.

### Domain-Specific Information:
- **Private Information Types:** Names (e.g., Mr. X, Ms. Y), dates,
addresses, and other identifying information.
- **Contextual Considerations:** Consider the context of the document
and the constitutional rules provided to ensure comprehensive coverage.
- **Ethical Standards:** Follow general legal and ethical standards
for identifying potentially private information.
- **Common Legal Entities:** Names of countries, courts, and 
organizations are often considered private unless specified 
otherwise.
- **Dates:** Include all dates mentioned in the text, especially
those related to legal proceedings.
- **Addresses:** If addresses are provided, they should be marked 
as private.
- **Names and Titles:** Always mark names and titles (e.g., Mr.,
Mrs., Ms.) as potentially private.
- **Legal Context:** Consider legal contexts such as filings, 
court dates, and proceedings.
- **Specific Examples:** Pay attention to specific examples provided
in the constitutional rules and extrapolate similarly.

### Example Strategies:
1. **Name Identification:** If a name is mentioned, mark it as 
private, even if it is not explicitly labeled as such in the 
rules.
2. **Date Annotation:** Mark all dates as private, regardless 
of context.
3. **Address Recognition:** If an address is provided, annotate
it as private.
4. **Legal Context:** Consider legal contexts such as filings, 
court dates, and proceedings.
5. **Extrapolation:** Use examples to extrapolate and cover 
similar patterns in the text.
6. **Ethical Standards:** Apply general legal and ethical standards
to identify private information.

### Final Instructions:
Follow the guidelines and use the provided strategies 
to systematically
annotate private phrases in the given text. Ensure comprehensive 
coverage and prioritize recall over precision. Use the feedback
to continually refine your approach to improve the accuracy 
of your annotations.

\end{verbatim}
\label{app:gepa_prompt_7b_contd}
\end{tcolorbox}
\end{table*}

\begin{table*}[t]
\centering
\caption{}
\begin{tcolorbox}[
  width=\textwidth,
  colback=white!95!gray,
  colframe=gray!80!black,
  title=GEPA learnt prompt using a 7B model for private-token tagging on the Legal dataset (Continued...),
  coltitle=white,
  fonttitle=\bfseries,
  colbacktitle=gray!80!black,
  enhanced,
  sharp corners=south
]
\begin{verbatim}

### Example Inputs and Outputs:
----------------------------
EXAMPLES REMOVED FOR BREVITY
----------------------------
### Additional Notes:

- Pay attention to names, dates, addresses, and other identifying information
that could be considered private.
- Consider the context of the document and the constitutional rules provided
to ensure comprehensive coverage.
- Use general legal and ethical standards if constitutional rules are not 
specified.

### Task:
Based on the provided examples and feedback, you need to develop a systematic
approach to annotate private phrases from the given text using the 
constitutional rules. Ensure that you mark all potentially private information
and prioritize recall over precision. Use the feedback to continually refine
your approach.

### Specific Strategies:

1. **Name Identification:** If a name is mentioned, mark it as private, 
even if it is not explicitly labeled as such in the rules.
2. **Date Annotation:** Mark all dates as private, regardless of context.
3. **Address Recognition:** If an address is provided, annotate it as private.
4. **Legal Context:** Consider legal contexts such as filings, court dates, 
and proceedings.
5. **Extrapolation:** Use examples to extrapolate and cover similar patterns
in the text.
6. **Ethical Standards:** Apply general legal and ethical standards to identify
private information.

### Continuous Improvement:

- Use the feedback to adjust and refine your approach to improve precision 
and recall.
- Focus on missed items and refine your annotations accordingly.
\end{verbatim}
\label{app:gepa_prompt_7b_contd_2}
\end{tcolorbox}
\end{table*}

\begin{table*}[t]
\centering
\caption{}
\begin{tcolorbox}[
  width=\textwidth,
  colback=white!95!gray,
  colframe=gray!80!black,
  title=GEPA learnt prompt using a 14B model for private-token tagging on the Legal dataset,
  coltitle=white,
  fonttitle=\bfseries,
  colbacktitle=gray!80!black,
  enhanced,
  sharp corners=south
]
\begin{verbatim}

INSTRUCTIONS

Your task involves analyzing legal documents from international human
rights tribunals. The documents contain personal and procedural
details of cases involving individuals, including personal names,
ages, addresses, dates, and legal actions taken. You need to
meticulously identify and mark all phrases that contain personally
identifiable information (PII) or sensitive details about the
individuals involved in these cases.

Given the absence of clear “CONSTITUTION RULES”, you should use
general principles of privacy and confidentiality. Err on the side
of caution to avoid false negatives (missing private pieces of
information).

For each text, produce a JSON object with a key “private_phrases”
whose value is an array of strings. Each string should represent a
phrase from the text that contains information that should be
treated as private or sensitive according to your judgment or
provided guidelines.

To optimize the F1 score, balance minimizing false positives
(incorrectly marking information as private) and false negatives
(overlooking truly private information). Aim to capture all
potential private information, even if it leads to some false
positives.

Identify a variety of information types including names of people
(lawyers, government officials, victims, defendants), dates (of
arrests, trials, hearings), places (cities, prisons, courts,
hospitals), organizations (legal, medical, governmental), and other
data that could be used to identify individuals or reveal sensitive
personal information.

From past examples, recognize common patterns of private information
like references to injuries, medical conditions, financial details
(such as compensation amounts), and other sensitive personal data.

Example analysis demonstrates capturing detailed personal
information, legal proceedings descriptions, and locations of
incidents or residences.

---

\end{verbatim}
\label{app:gepa_prompt_14b}
\end{tcolorbox}
\end{table*}

\begin{table*}[t]
\centering
\caption{}
\begin{tcolorbox}[
  width=\textwidth,
  colback=white!95!gray,
  colframe=gray!80!black,
  title=GEPA learnt prompt using a 14B model for private-token tagging on the Legal dataset (Continued...),
  coltitle=white,
  fonttitle=\bfseries,
  colbacktitle=gray!80!black,
  enhanced,
  sharp corners=south
]
\begin{verbatim}

EXAMPLE FORMAT

clinical_text
TEXT TO ANALYZE HERE

output
private_phrases:

* phrase1
* phrase2
* phrase3

---

DOMAIN INFORMATION

PII and sensitive details in legal contexts include names,
birthdates, addresses, organization affiliations, specific legal
cases, and personal conditions or events.

Dates, locations, and organizations often hold significant
relevance and should be marked as private to protect individual
identities and confidentiality.

Ensure thorough identification of individuals involved in the
cases, including lawyers, defendants, victims, and officials.

Be cautious with medical details, financial amounts, and legal
statuses as these can also constitute sensitive personal
information.


\end{verbatim}
\label{app:gepa_prompt_14b_contd}
\end{tcolorbox}
\end{table*}

\subsection{Learnt prompts for MIPRO - Legal}
\label{app:mipro_prompts}
MIPRO-learned prompts for the Annotator Agent on the ECHR dataset are provided as follows: 3B model prompts in Table~\ref{app:mipro_prompt_3b}, 7B model prompts in Table~\ref{app:mipro_prompt_7b}, and 14B model prompts in Table~\ref{app:mipro_prompt_14b}.

\begin{table*}[t]
\centering
\caption{}
\begin{tcolorbox}[
  width=\textwidth,
  colback=white!95!gray,
  colframe=gray!80!black,
  title=MIPRO learnt prompt using a 3B model for private-token tagging on the Legal dataset,
  coltitle=white,
  fonttitle=\bfseries,
  colbacktitle=gray!80!black,
  enhanced,
  sharp corners=south
]
\begin{verbatim}
Annotate the provided clinical texts to identify private entities according
to the following systematic rules:

1. Names of people (full names or initials).
2. Locations (specific place names).
3. Addresses (full addresses).
4. Phone numbers.
5. Email addresses.
6. Dates (exact dates or ranges).
7. Medical records (such as diagnoses or treatments).
8. Financial information (such as bank account numbers, credit card details).
9. Any other information that, according to common sense, should not be 
disclosed publicly.

Please apply these rules to the provided clinical texts and annotate any 
matching private phrases in the following format:

```json
{
 "private_phrases": ["phrase1", "phrase2", "phrase3"]
}
```

Text:
\end{verbatim}
\label{app:mipro_prompt_3b}
\end{tcolorbox}
\end{table*}

\begin{table*}[t]
\centering
\caption{}
\begin{tcolorbox}[
  width=\textwidth,
  colback=white!95!gray,
  colframe=gray!80!black,
  title=MIPRO learnt prompt using a 7B model for private-token tagging on the Legal dataset,
  coltitle=white,
  fonttitle=\bfseries,
  colbacktitle=gray!80!black,
  enhanced,
  sharp corners=south
]
\begin{verbatim}
You are a constitutional privacy annotation agent. Your task is to extract
private phrases from the provided clinical text, ensuring high precision 
and recall. You will be given some rules in natural language in the form of
a constitution, use that to annotate private leaking stuff. Systematically 
apply the constitution rules to identify all private phrases. If no 
constitution is available, annotate based on your best judgment. When in
doubt, err on the side of marking something as private to avoid false 
negatives (FN). 

**ANNOTATION FORMAT:** 
```json
{
 "private_phrases": ["phrase1", "phrase2", "phrase3"]
}
```

Here are the **CONSTITUTION RULES** (if provided):

**ANNOTATION TASK:**
Apply the constitution rules systematically to extract all matching 
private phrases. Include all names, dates, places, and any other information
that could potentially identify individuals, dates, or locations.

Example Constitution Rules:
- Names of individuals
- Dates
- Places
- Case numbers
- Identification numbers

If the constitution does not explicitly mention a type of phrase, 
consider it private if it could be used to identify an individual or location.

Now, based on the provided rules, annotate the following text:

**Text:**
\end{verbatim}
\label{app:mipro_prompt_7b}
\end{tcolorbox}
\end{table*}

\begin{table*}[t]
\centering
\caption{}
\begin{tcolorbox}[
  width=\textwidth,
  colback=white!95!gray,
  colframe=gray!80!black,
  title=MIPRO learnt prompt using a 14B model for private-token tagging on the Legal dataset,
  coltitle=white,
  fonttitle=\bfseries,
  colbacktitle=gray!80!black,
  enhanced,
  sharp corners=south
]
\begin{verbatim}
Given a piece of clinical or procedural text related to legal cases involving
human rights issues, your task is to extract all private phrases that could
potentially reveal sensitive information, such as names, addresses, dates, 
and other identifiers. Your goal is to ensure a comprehensive extraction with
minimal false negatives, even if it means risking false positives. Below is 
the provided text. Please adhere to the following guidelines while extracting
private phrases:

- Names of individuals (first name + last name, or last name alone if commonly
used in the document).
- Legal representatives and their professional titles.
- Organizations, institutions, and official bodies mentioned.
- Addresses and locations.
- Dates mentioned, especially those tied to specific events or milestones.
- Any unique identifiers or codes associated with legal documents or processes.

Apply these principles systematically to extract all potential private phrases
from the text. The output should be formatted as a JSON object, listing each 
private phrase individually.

**EXAMPLE OUTPUT FORMAT**

```json
{
 "private_phrases": ["phrase1", "phrase2", "phrase3"]
}
```

**TEXT FOR ANNOTATION**
[Provide the clinical/procedural text here]

Please apply the aforementioned instructions to the provided text to 
generate the most accurate set of private phrases possible.
\end{verbatim}
\label{app:mipro_prompt_14b}
\end{tcolorbox}
\end{table*}

\subsection{LLM-as-a-judge prompt for computing agent performance}
\label{app:llm_as_a_judge}
The full LLM-as-a-judge prompt used to compare MAC+ performance is shown in Tables~\ref{app:llm_judge_1} and \ref{app:llm_judge_2}.

\begin{table*}[t]
\centering
\caption{}
\begin{tcolorbox}[
  width=\textwidth,
  colback=purple!6!white,
  colframe=purple!45!black,
  title=LLM-as-a-Judge Prompt for Evaluating Agent Performance,
  coltitle=white,
  fonttitle=\bfseries,
  colbacktitle=purple!45!black,
  enhanced,
  sharp corners=south
]
\begin{verbatim}
SYSTEM: You are an **LLM-as-a-judge**.

Your job is to:

1. **Identify the task** based solely on the provided input.
2. You will then be given **two candidate outputs** for this task:

   * One from **Model A**
   * One from **Model B**
3. Your task is to **select which output is better**: **A** or **B**.

### Selection Criteria

Choose the better output based on:

* **Quality of the answer**
* **Clarity and correctness of the explanation**

### Task Context

All input tasks relate to **updating a constitution** that is used by an agent 
to **identify and mark private tokens (PII)**. The task will be one of the 
following types:

1. Decide whether to **add, remove, or edit** rules based on the current 
constitutional state and observed errors.
2. Receive an **index of an existing rule** and **update it** by writing a
**high-quality revised rule** based on the provided reasoning.
3. **Create a new rule** from provided guidance and write a **high-quality
rule**.

### Output Format

You must output **only a JSON 
object** in the following format:

```json
{
  "selection": "A"
}
```

or

```json
{
  "selection": "B"
}
```

\end{verbatim}
\label{app:llm_judge_1}
\end{tcolorbox}
\end{table*}

\begin{table*}[t]
\centering
\caption{}
\begin{tcolorbox}[
  width=\textwidth,
  colback=purple!6!white,
  colframe=purple!45!black,
  title=LLM-as-a-Judge Prompt for Evaluating Agent Performance,
  coltitle=white,
  fonttitle=\bfseries,
  colbacktitle=purple!45!black,
  enhanced,
  sharp corners=south
]
\begin{verbatim}

USER: Here is the definition of the task, and the sample input. Note, it is a 
prompt, just understand this task but do not start doing the task itself that
is not ure job. You should just use this to understand the task, that's it.

**YOU MUST JUDGE BASED ONLY ON THE INPUT.**

```
{{INPUT}}
```

USER: Here is the output from Model A based on the input:

```
{{OUTPUT A}}
```

USER: Here is the output from Model B based on the input:

```
{{OUTPUT B}}
```

USER: Now, based on the task described in the input and the two outputs above,
select **A** or **B**.


\end{verbatim}
\label{app:llm_judge_2}
\end{tcolorbox}
\end{table*}

\end{document}